\def\year{2021}\relax

\documentclass[letterpaper]{article}
\usepackage{aaai21}  
\usepackage{times}  
\usepackage{helvet} 
\usepackage{courier}  
\usepackage[hyphens]{url}  
\usepackage{graphicx} 
\usepackage{natbib}  
\usepackage{caption} 
\usepackage[many]{tcolorbox}
\usepackage{amsmath,amssymb, graphicx}
\usepackage[ruled,linesnumbered]{algorithm2e}
\usepackage{algorithmic}
\usepackage{paralist} 
\usepackage{color}
\usepackage{multirow}
\usepackage{rotating}
\usepackage{todonotes}
\usepackage[mathscr]{eucal} 
\usepackage{amsbsy} 
\usepackage{subfigure}
\usepackage{booktabs}
\usepackage{xcolor}
\usepackage{tikz}
\usepackage{multirow}
\usepackage[mathscr]{eucal} 
\usepackage{empheq} 
\usepackage{epstopdf}
\usepackage{balance}
\usepackage{empheq} 
\renewcommand{\algorithmicrequire}{\textbf{Input:}}

\newcommand{\docvec}{\texttt{Doc2Vec}\xspace}
\newcommand{\VizFake}{\texttt{VizFake}\xspace}
\newcommand{\fasttext}{\texttt{FastText}\xspace}
\newcommand{\tfidf}{\texttt{TF-IDF}\xspace}
\newcommand{\GloVeLSTM}{\texttt{GloVe\slash LSTM}\xspace}

\usepackage{moresize}
\usepackage{tabularx}
\usepackage{enumitem}
\setlist{nolistsep}

\newcommand{\hide}[1]{}

\usepackage{url}

\renewcommand{\algorithmicrequire}{\textbf{Input:}}

\newcounter{ALC@tempcntr}

\begin{document}
 \author{
    {Sara Abdali\textsuperscript{\rm 1}}, {Rutuja Gurav\textsuperscript{\rm 1}}, {Siddharth Menon\textsuperscript{\rm 1}}, {Daniel Fonseca\textsuperscript{\rm 1}},
 { \textbf{Negin Entezari\textsuperscript{\rm 1}}},
 { \textbf{Neil Shah\textsuperscript{\rm 2}}}, {\textbf{Evangelos E. Papalexakis
 \textsuperscript{1}}}

}
\affiliations{
    \textsuperscript{\rm 1}
    Department of Computer Science and Engineering University of California, Riverside\\ 900 University Avenue, Riverside, CA, USA\\
    \textsuperscript{ \rm 2}{ Snap Inc.}\\
    \{sabda005,rgura001,smeno004,dfons007,nente001\}@ucr.edu\\
 epapalex@cs.ucr.edu\\
 nshah@snap.com \\

}
\title{Identifying Misinformation from Website Screenshots}
\maketitle
\begin{abstract}
Can the look and the feel of a website give information about the trustworthiness of an article? In this paper, we propose to use a promising, yet neglected aspect in detecting the misinformativeness: the overall look of the domain webpage. To capture this overall look, we take screenshots of news articles served by either misinformative or trustworthy web domains and leverage a tensor decomposition based semi-supervised classification technique.
The proposed approach i.e., \VizFake is insensitive to a number of image transformations such as converting the image to grayscale, vectorizing the image and losing some parts of the screenshots. \VizFake leverages a very small amount of known labels, mirroring realistic and practical scenarios, where labels (especially for known misinformative articles), are scarce and quickly become dated. The F1 score of \VizFake on a dataset of 50k screenshots of news articles spanning more than 500 domains is roughly 85\% using  only 5\% of ground truth labels. Furthermore, tensor representations of \VizFake, obtained in an unsupervised manner, allow for exploratory analysis of the data that provides valuable insights into the problem.
Finally, we compare \VizFake with deep transfer learning, since it is a very popular black-box approach for image classification and also well-known text text-based methods. \VizFake achieves competitive accuracy with deep transfer learning models while being two orders of magnitude faster and not requiring laborious hyper-parameter tuning.
\end{abstract}
\begin{figure}[tb]
    \begin{center}
  \includegraphics[width = 1\linewidth]{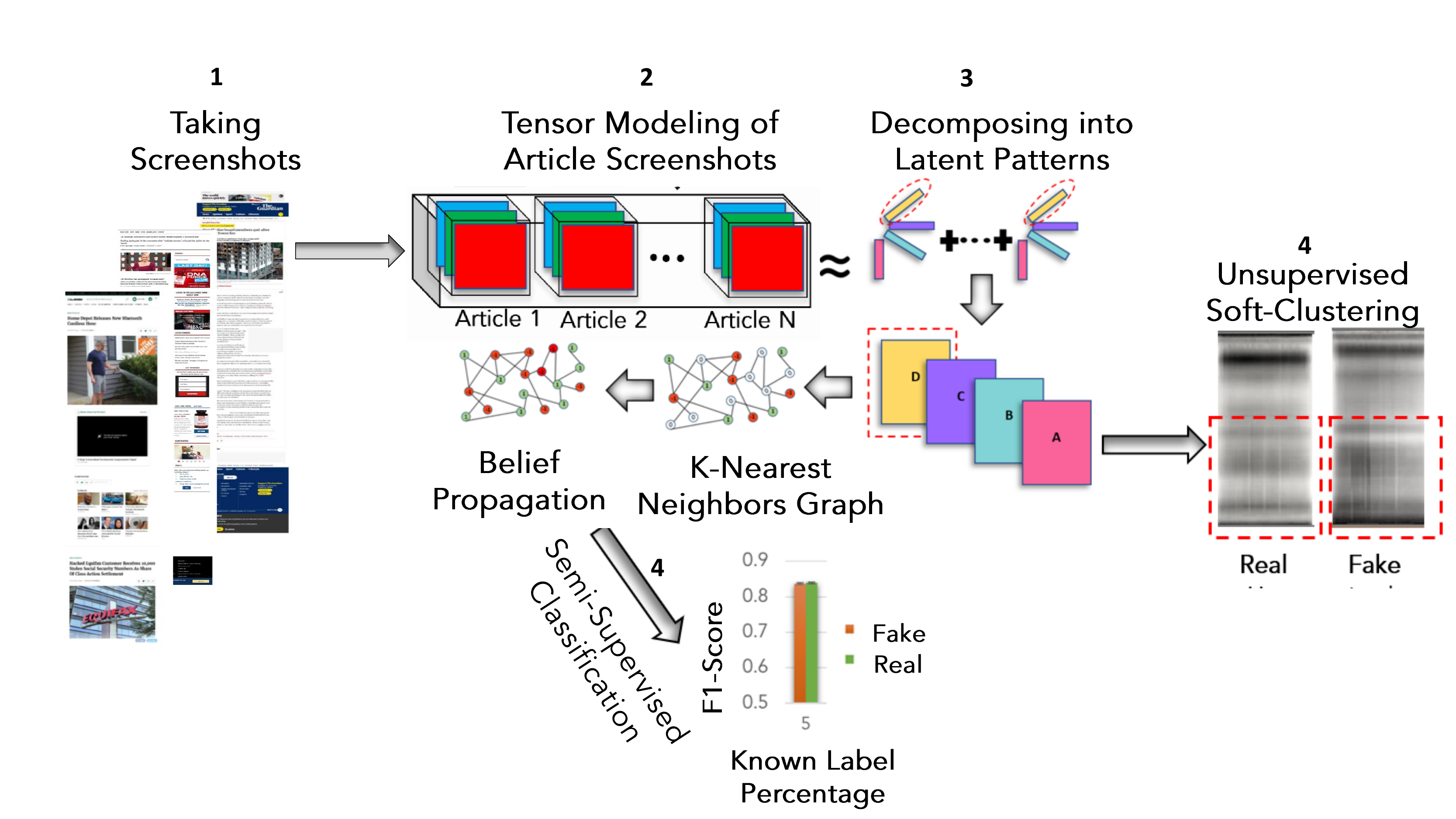}
    \end{center}
    \caption{Creating a tensor-based model out of news articles' screenshots and decomposing the tensor using CP/PARAFAC into latent factors and then creating a nearest neighbor graph based on the similarity of latent patterns and leveraging belief propagation to propagate very few known labels throughout the graph. As illustrated, the F1 score of both real and fake classes is roughly 85\% using just 5\% of known labels. Moreover, VizFake has exploratory capabilities for unsupervised clustering of screenshots.}
    \label{fig:Crown-Jewel}
\end{figure}
\section{Introduction}
\label{sec:intro}
Despite the benefits that the emergence of web-based technologies has created for news and information spread, the increasing spread of fake news and misinformation due to access and public dissemination functionalities of these technologies has become increasingly apparent in recent years. Given the growing importance of the fake news detection task on web-based outlets, researchers have placed considerable effort into design and implementation of efficient methods for finding misinformation on the web, most notably via natural language processing methods \cite{ciampaglia2015computational,Rubin2016,Horne2017,shu2017fake}. intended to discover misinformation via nuances in article text. article's text Although utilizing textual information is a natural approach, there are few drawbacks: most notably, such approaches require complicated and time-consuming analysis to extract linguistic, lexical, or psychological features such as sentiment, entity usage, phrasing, stance, knowledge-base grounding, etc. Moreover, the problem of identifying misinformativeness using textual cues is challenging to define well, given that each article is composed of many dependent statements (not all of which are fact-based) and editorialization. Finally, most such approaches require extremely large labeled sets of misinformative articles, which are often unavailable in practice due to lack of reliable human annotators, as well as quickly become ``dated'' due to shift in topics, sentiment, and reality and time itself. These article-based labels inherently result in event-specificity and bias in resulting models, which can lead to poor generalization in the future for different article types.

In this work, we take a step back to tackle the problem with a human, rather than an algorithmic perspective. We make two choices that are not made jointly in prior work. Firstly, we tackle misinformation detection by leveraging a \emph{domain-level} feature. Secondly, we focus on the discovery of misinformation using \emph{visual cues} rather than textual ones. We expand upon these two points below.
Firstly, leveraging domain features for misinformation detection is not only an easier but also a likely more fruitful/applicable problem setting in practice. In reality, most highly reputed news sources do not report misinformative articles due to high editorial standards, scrutiny, and expectations. For example, the public fallout from misinformation being spread through famous organizations like CNN or BBC would be disastrous. However, there are many misinformation farms and third-parties which create new domains with the intent of deceiving the public \cite{boatwright2018troll}. Moreover, these actors have little incentive to spread real articles in addition to fake ones. Thus, in most cases, domain feature could prove to be a better target to stymie the spread of misinformation. Conveniently, several crowd-sourced tools and fact-checkers like BS Detector \footnote{http://bsdetector.tech/} or Newsguard \footnote{https://www.newsguardtech.com} provide domain-level labels rather than article-level, which we utilize here. 

Secondly, visual cues are a promising, yet underserved research area, especially in the context of misinformation detection. While past literature in text-based methods in this space is rich (see \cite{oshikawa2018survey} for an overview), prior work on visual cues is sparse.  Past works \cite{visual2017,fakeimages,rumors-imagetext} primarily focus on doctored/fake-news associated images and visual coherence of images with article text.  However, since these works are limited to fake news which spreads with images, they are inapplicable for articles which do not incorporate multimedia. Moreover, these works all have inherent article specificity, and none consider the overall visual look and representation of the hosting domain or website for a given article.  Intuitively and anecdotally, in contrast to unreliable sources that tend to be visually messy and full of advertisements and popups, trustworthy domains often look professional and ordered.  For example, real domains often request users to agree to privacy policies, have login/signup/subscription functionalities, have multiple featured news articles clearly visible, etc.  Conversely, strong tells for fake domains tend to include errors, negative space, unprofessional/hard-to-read fonts, and blog-post style \cite{Website_Design,formality_matter,Website_Quality}. Figure \ref{fig:Crown-Jewel} demonstrates this dichotomy with a few examples. While we as humans use these signals to quickly discern the quality and reliability of news sources without delving into the depth of the text, prior works have not directly considered them. Thus, we focus on bridging this gap with the assumption that many misinformative articles do not need to be read to be suspected. 

Given these two facets, we ask: {\it ``can we identify misinformation by leveraging the visual characteristics of their domains?''}  In this work, we propose an approach for classification of article screenshots using image processing approaches.  In contrast to deep learning approaches such as convolutional neural networks (CNNs) which take a relatively long time to train, are data-hungry, and require careful hyperparameter tuning, we propose a novel tensor-based semi-supervised classification approach which is fast, efficient, robust to image resolution, and missing image segments, and data-limited.  We demonstrate that our approach henceforth refereed to as \VizFake,  can successfully classify articles into fake or real classes with an F1 score of 85\% using very few (i.e., $<$ 5\% of available labels).  Summarily, our major contributions are as follows:
\begin{itemize}
\item {\bf Using visual signal for modeling domain structure}: 
We propose to model article screenshots from different domains using a tensor-based formulation.
\item {\bf Fast and robust tensor decomposition
approach for classification of visual information}:
We propose a tensor-based model to find latent article patterns. We compare it against typical deep learning models. \VizFake performs on par, while being significantly faster and needless to laborious hyperparameter tuning.
\item {\bf Unsupervised exploratory analysis}: Tensor-based representations of \VizFake derived in an unsupervised manner, allow for interpretable exploratory analysis of the data which correlate with existing ground truth.
\item {\bf Performance in label-scarce settings}:
In contrast to deep learning approaches, \VizFake is able to classify news articles with high performance using very few labels, due to a semi-supervised belief propagation formulation. 
\item {\bf Experimenting on real-world data}:
We evaluate \VizFake on a real-world dataset we constructed with over 50K news article screenshots from more than 500 domains, by extracting tweets with news article links. Our experiments suggest strong classification results (85\% F1 score) with very few labels ($<$ 5\%) and over two orders of speedup compared to CNN-based methods.
\end{itemize}
\par The remainder of this paper is organized as follows:  First, the proposed \VizFake is described. Next, we discuss the implementation details and the dataset. Afterwards, the experimental evaluation of the \VizFake as well as variants and baselines is presented. Then, we discuss the related work, and finally we draw the conclusions.

\section{Proposed Method}
\label{sec:method}

Here, we discuss our formulation and proposed semi-supervised tensor-based approach i.e., \VizFake method.
\subsection{Problem Formulation}
\label{sec:problemform}
We solve the following problem:
\tcbset{colback=orange!5!white,colframe=orange!30!white}

\begin{tcolorbox}[width=1\linewidth]
{\bf Given} (i) a collection of news domains and a number of full-page screenshots of news articles published by each domain and (ii) a small number of labels. 

{\bf Classify} the unlabeled screenshots as misinformation or not.
 \end{tcolorbox}

\subsection{Semi-supervised Tensor-based Method i.e VizFake}
\par  \VizFake aims to explore the predictive power of  visual information about articles published by domains. As we argued above, there is empirical evidence that suggests this proposition is plausible.Thus, we introduce a novel model to leverage this visual information.
We propose a tensor-based semi-supervised approach that is able to effectively extract and use the visual cue which yields highly predictive representations of screenshots, even with limited supervision, also, due to its elegant and simple nature, allows for interpretable exploration. \VizFake has the following steps:

\subsubsection{Tensor-based modeling}
The first step of \VizFake refers to constructing a tensor-based model out of articles' screenshots.
RGB digital images are made of pixels each of which is represented by three channels, i.e., red, green, and blue. So each image channel shows the intensity of the corresponding color for each pixel of the image.

A tensor is a multi-way array. We use a 4-mode tensor embedding for modeling news articles' screenshots. since each channel of an RGB digital image is a matrix, by stacking all three channels we create a 3-mode tensor for each screenshot and if we put all 3-mode tensor together, we create a 4-mode tensor out of all screenshots. 

\subsubsection{Tensor Decomposition}
 As we mentioned above, a tensor is a multi-way array, i.e., an array with three or more dimensions. The Canonical Polyadic (CP) or PARAFAC decomposition, factorizes a tensor into a summation of $R$ rank-one tensors. For instance, a 4-mode tensor $\boldsymbol{\mathcal{X}}$ of dimensions $I\times J\times K \times L$ is decomposed into a sum of outer products of four vectors as follows:$$
\boldsymbol{\mathcal{X}} \approx \Sigma_{r=1}^{R} \mathbf{a}_r \circ \mathbf{b}_r \circ \mathbf{c}_r  \circ \mathbf{d}_r
$$
where $\mathbf{a}_r \in \mathbb{R}^{I}$, $\mathbf{b}_r \in \mathbb{R}^{J}$, $\mathbf{c}_r \in \mathbb{R}^{K}$ $\mathbf{d}_r \in \mathbb{R}^{l}$  
and the outer product is given by \cite{Papalexakis:2016,Tensor}: 
$$
( \mathbf{a}_r , \mathbf{b}_r , \mathbf{c}_r , \mathbf{d}_r) (i,j,k,l)=\mathbf{a}_r(i) \mathbf{b}_r(j) \mathbf{c}_r(k) \mathbf{d}_r(l)  \forall{i,j,k,l}
$$

We define the factor matrices as $\mathbf{A} = [\mathbf{a}_1 ~ \mathbf{a}_2 \ldots \mathbf{a}_R]$, $\mathbf{B} = [\mathbf{b}_1 ~ \mathbf{b}_2 \ldots \mathbf{b}_R]$, $\mathbf{C} = [\mathbf{c}_1 ~ \mathbf{c}_2 \ldots \mathbf{c}_R]$ and $\mathbf{D} = [\mathbf{d}_1 ~ \mathbf{d}_2 \ldots \mathbf{d}_R]$
where $\mathbf{A} \in \mathbb{R}^{I\times R}$, $\mathbf{B} \in \mathbb{R}^{J  \times R}$, $\mathbf{C} \in \mathbb{R}^{K \times R}$ and $\mathbf{D} \in \mathbb{R}^{L \times R}$ denote the factor matrices and $R$ is the rank of decomposition or the number of columns in the factor matrices. Moreover, the optimization problem for estimating the factor matrices is defined as follows:
 
$$
 \min_{\mathbf{A,B,C,D}} ={{\lVert} \boldsymbol{\mathcal{X}} - \Sigma_{r=1}^{R} \mathbf{a}_r \circ \mathbf{b}_r \circ \mathbf{c}_r \circ \mathbf{d}_r{\rVert}}^2
$$
\par For solving the optimization problem above we use Alternating Least Squares (ALS) which solves for any of the factor matrices by fixing the others due to simplicity and the speed of this algorithm \cite{Papalexakis:2016,Tensor}.
\par Having the mathematical explanation above in mind, the second step of \VizFake is decomposition of the proposed tensor-based model for finding the factor matrix corresponding to article mode, i.e., factor matrix $\mathbf{D}$ which comprises latent patterns of screenshots. We will leverage these latent patterns for screenshot classification.

 \begin{figure}[!t]
    \begin{center}
    \includegraphics[width = 1\linewidth]{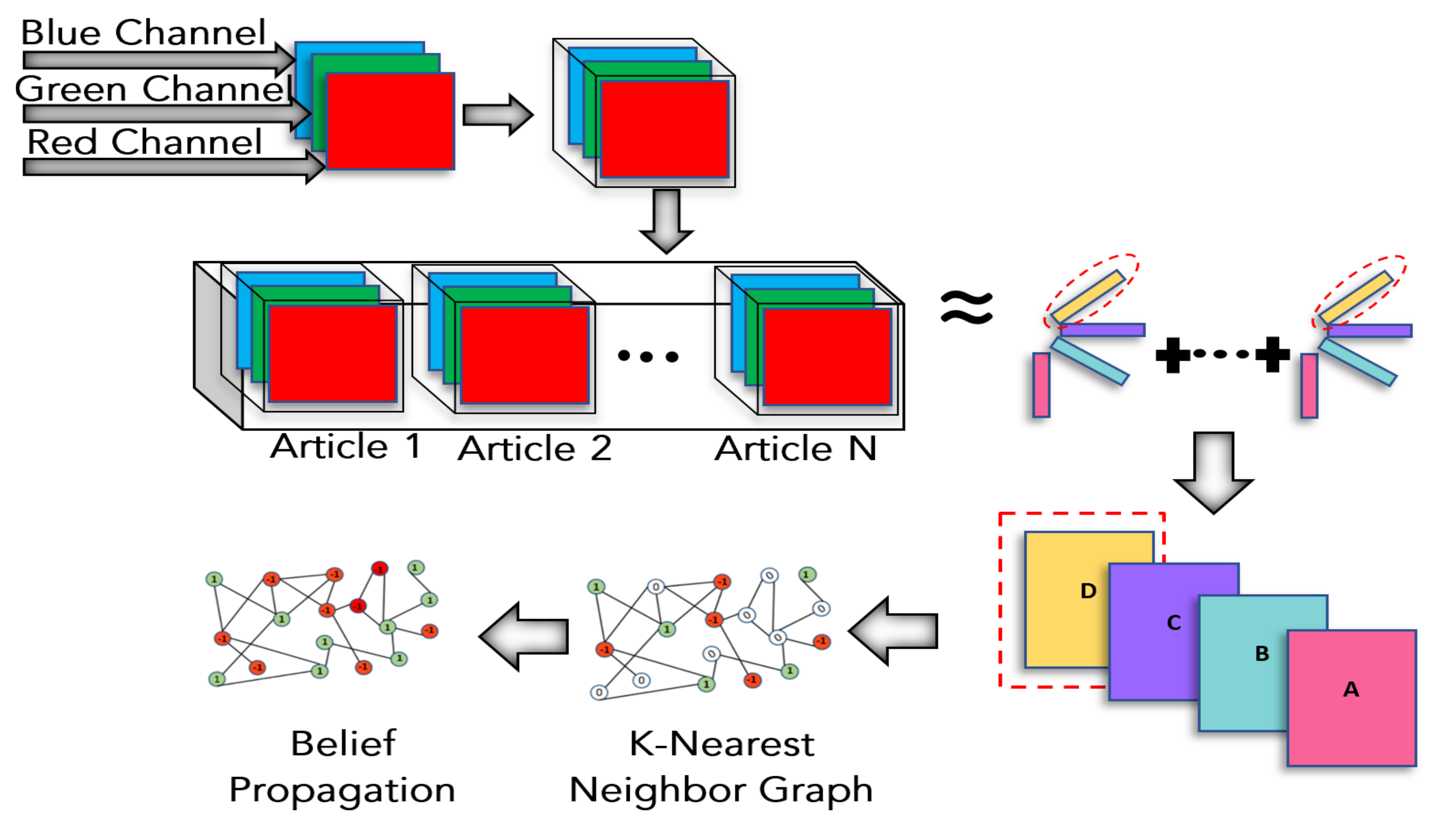}
    \end{center}
    \caption{Proposed tensor-based modeling and semi-supervised classification of the screenshots i.e. VizFake.}
    \label{fig:decomposition}
\end{figure}.
\subsubsection{Semi-supervised classification}
The third and last step of \VizFake is the classification of news articles using the factor matrix $\mathbf{D}$ corresponding to article mode resulted from the decomposition of the tensor-based model.
\par As we mentioned before, each factor matrix comprises the latent patterns of the corresponding mode in $\mathbf{R}$ dimensional space. Therefore, each row of factor matrix $\mathbf{D}$ is an $\mathbf{R}$ dimensional representation of the corresponding screenshot. So, we can consider each screenshot as a data point in $\mathbf{R}$ dimensional space. We create a K-nearest neighbor graph (K-NN) Graph by considering data points as nodes, and the Euclidean distance between the nodes as edges of the graph.
\par Belief propagation \cite{SurveyBP,BP} is a message passing-based algorithm that is usually used for calculating the marginal distribution on graph-based models such as Bayesian networks, Markov, or K-NN graph. In this algorithm, each node of a given graph leverages the messages received from neighboring nodes to compute its belief (label) using the following iterative update rule:
$$
b_i(x_i)\propto \prod_{j\in N_i}m_{j\hookrightarrow i}(x_i) 
$$
where $b_i(x_i)$ denotes the belief of node $i$, $m_{j\hookrightarrow i}(x_i)$ is a message sent from node $j$ to node $i$ and conveys the opinion of node $j$ about the belief of node $i$, and $N_i$ denotes all the neighboring nodes of node $i$ \cite{SurveyBP,BP}.
 \par Since we model homophily (similarity) of screenshots patterns using a K-NN graph as explained above, we can leverage Belief Propagation in a semi-supervised manner to propagate very few available labels throughout the graph. 
A fast and linearized implementation of Belief propagation is proposed in \cite{KoutraKKCPF11} which solves the following linear system: 
$$[\mathbf{I} + a \mathbf{D} - c'\mathbf{A}]b_h = \phi_h$$
where $\phi_h$ and $b_h$ stand for the prior and the final beliefs, respectively. $\mathbf{A}$ denotes the $n \times n$ adjacency matrix of the K-NN graph, $\mathbf{I}$ denotes the $n \times n$ identity matrix, and $\mathbf{D}$ is a $n \times n$  matrix where $\mathbf {D}_{ii}= \sum_j \mathbf{A}_{ij}$ and $\mathbf{D}_{ij}=0$ for $i \neq j$. $a$ and $c^{'}$ are also defined as:
 $a=\frac{4h_h^2}{1-4h_h^2}$ , $c'= \frac{2h_h}{(1-4h_h^2)}$
where $h_h$ denotes the homophily factor between nodes. In fact, higher homophily corresponds to having more similar labels. Readers are referred to \cite{KoutraKKCPF11} for more details. An overview of \VizFake is depicted in Figure \ref{fig:decomposition}.

\section{Experimental Evaluation}
\label{sec:experiments}

In this section, we first discuss implementation and dataset details and then report a set of experiments to investigate the effect of changing rank, resolution, and some image transformation on the performance of \VizFake and then we compare it against CNN deep-learning model,  text-base text-based approaches and webpage structure features.
\subsection{Dataset Description}
Although collecting human annotation for misinformation detection is a complicated and time-consuming task, there exist some crowd-sourced schemes such as the browser extension ``BS Detector'' which provide a number of label options, allowing users to label domains into different categories such as biased, clickbait, conspiracy, fake, hate, junk science, rumor, satire, unreliable, and real.  We use BS Detector as our ground truth and consider all of the nine categories above  but ``real'' as ``fake'' class. We reserve a more fine-grained analysis of different ``fake'' categories for future work (henceforth collectively refer to all of those categories of misinformation as ``fake''). 

\par We describe our crawling process  in order to promote reproducibility, as we are unable to share the data because of copyright considerations. We crawled Twitter to create a dataset out of tweets published between June and August 2017 which included links to news articles. Then, we implemented a javascript code using Node.js open source server environment and Puppeteer library for automatically taking screenshots of scrolled news articles of our collected dataset.
\begin{itemize}
\item we took screenshots of 50K news articles equally from more than 500 fake and real domains i.e., a balanced dataset including 50\% from fake and 50\% real domains.
\item To investigate the effect of class imbalance, we created an imbalanced dataset of the same size, i.e., 50k but this time we selected $\frac{2}{3}$ of the screenshots from real domains and $\frac{1}{3}$ of the data from fake ones.
\item Although we tried to select an equal number of articles per each domain, sometimes fake domains do not last long and the number of fake articles published by them is limited. However, we show that this limitation does not affect the classification, because the result of the fake discrimination is almost same as the real class.
\end{itemize}
\begin{figure}[!tb]
    \begin{center}
\subfigure[F1-Balanced]{\includegraphics[width = 0.495\linewidth]{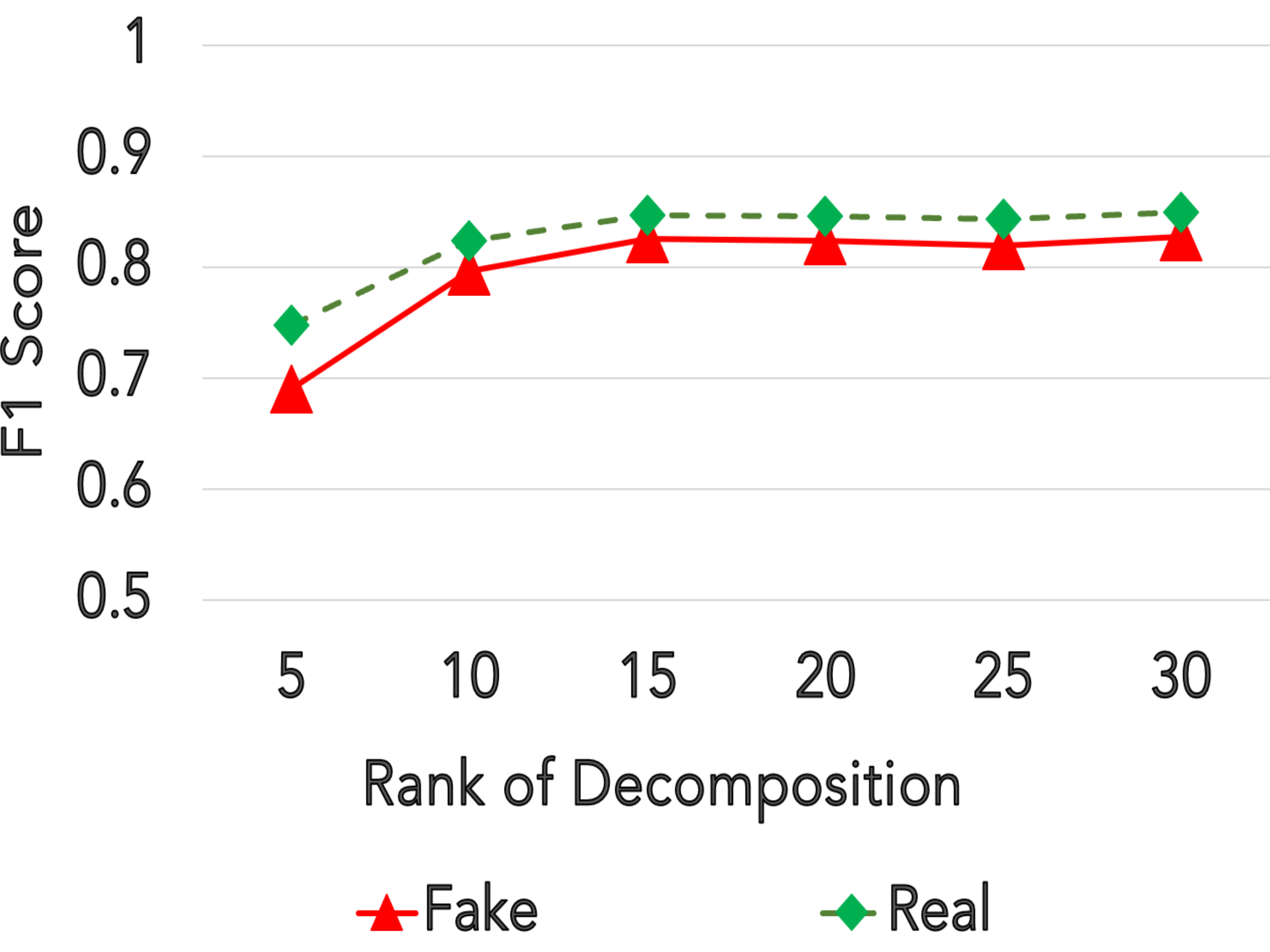}}
\subfigure[F1-Imbalanced]{\includegraphics[width = 0.495\linewidth]{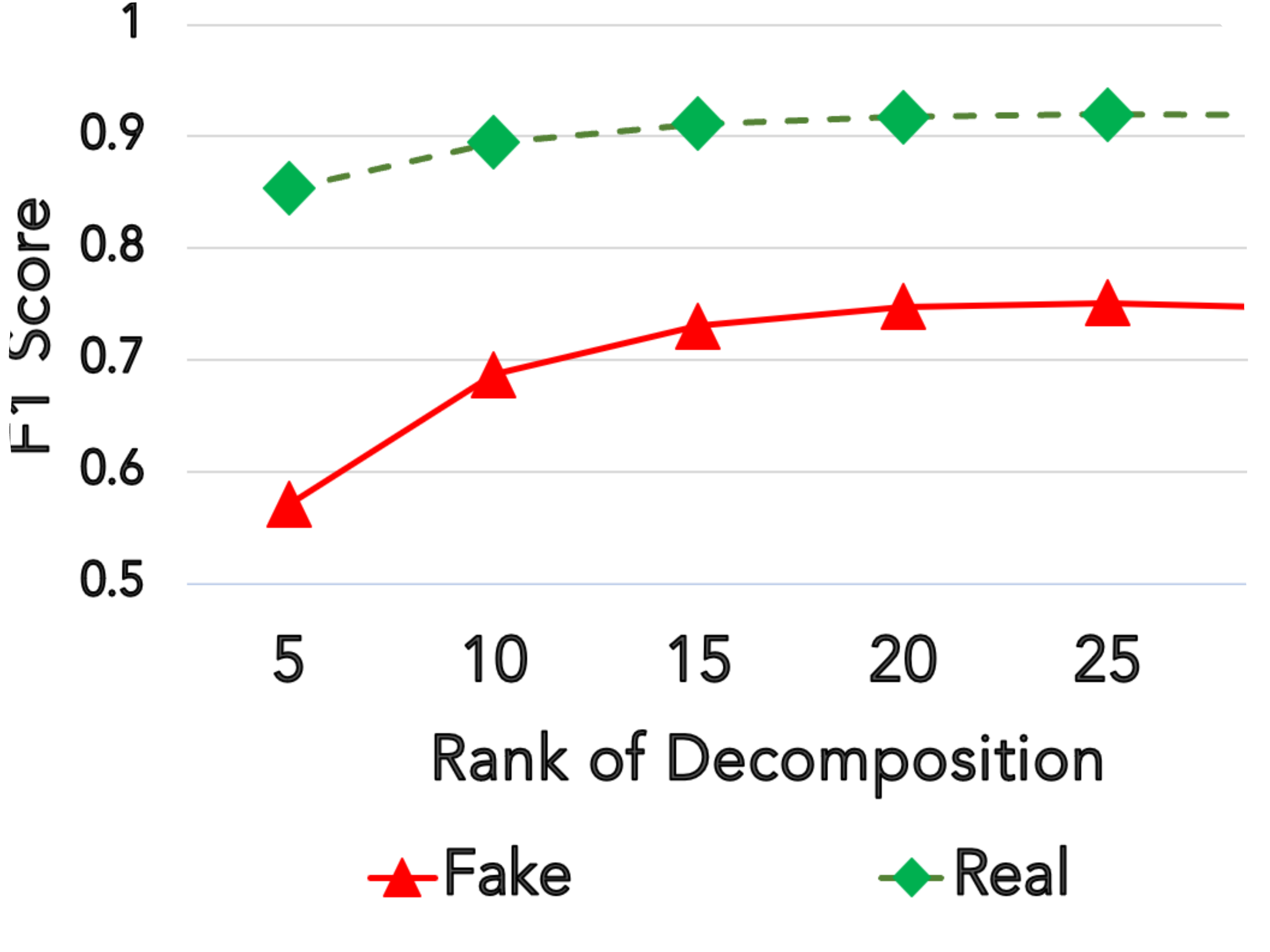}}
    \end{center}
    \caption{F1 score of VizFake for different ranks when experimenting on balanced and imbalanced datasets. The best ranks for these datasets are 15 and 25 respectively.}
    \label{fig:rank}
\end{figure}
\subsection{Implementation Details}
\par We used Matlab for implementing \VizFake approach and for CP/PARAFAC decomposition we used Tensor Toolbox version 2.6 \footnote{\tiny {https://www.sandia.gov/~tgkolda/TensorToolbox/index-2.6.html}}\cite{TTB_Dense}. For Belief Propagation, we used Fast Belief Propagation (FaBP) \cite{KoutraKKCPF11} which is linear in the number of edges. For finding the best rank of decomposition $R$ and the number of nearest neighbors $K$ for both balanced and imbalanced datasets, we grid searched the values between range 5-30 for $R$ and 1-50 for $K$. Based on our experiments, we set $R$ to 15 and 25 for balanced and imbalanced datasets, respectively and set $K$ to 20 for both datasets. We measured the effectiveness of \VizFake using widely used F1 score, precision, and recall metrics. We run all of the experiments 25 times and we report the average and standard deviation of the results for all mentioned metrics. The F1 score of different ranks for balanced and imbalanced datasets and both real and fake classes is shown in Figure \ref{fig:rank}.

\begin{figure}[!ht]
    \begin{center}
\subfigure[F1-Real]{\includegraphics[width = 0.65\linewidth]{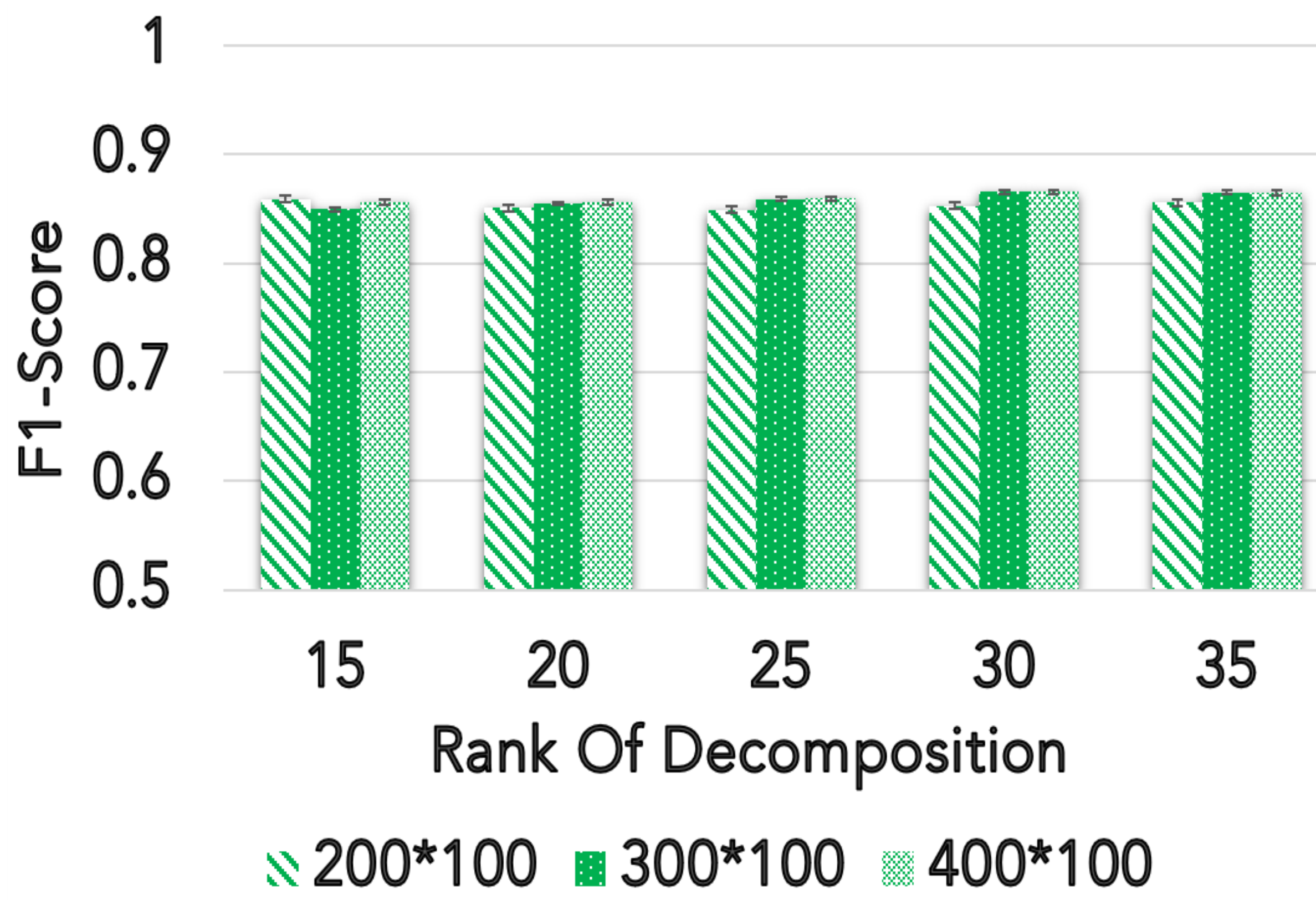}}
\subfigure[F1-Fake]{\includegraphics[width = 0.7\linewidth]{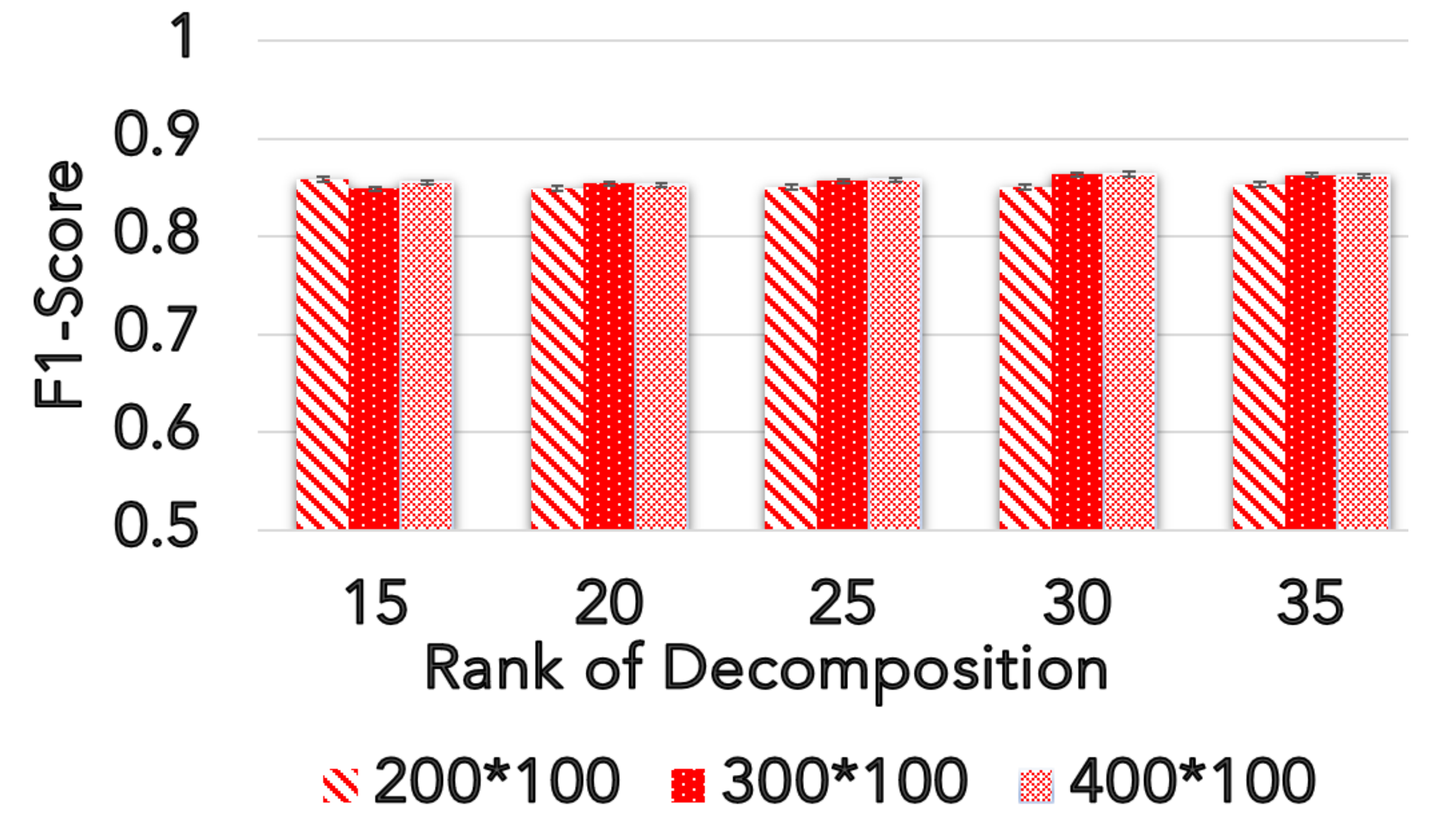}}
    \end{center}
    \caption{F1 score of VizFake for different resolutions. F1 score increases slightly when experimenting on higher resolution images.}
    \label{fig:resolution}
\end{figure}

\subsection{Investigating Detection Performance}

First, we aim at investigating the detection performance of \VizFake in discovering misinformative articles.  
A caveat in experimentation is that different articles even from the same domains may have different lengths, and thus screenshots of a fixed resolution may capture more or less information from different articles.  However, fixed-resolution is an important prerogative for \VizFake (and many others), thus we must use the same length for all screenshots.  
\par Thus, we first evaluate the effect of resolution to choose a fixed setting for our model in further experiments. We experiment on screenshots of size $200 \times 100$, $300\times100$, and $400\times100$, and simultaneously evaluate the effect of different decomposition rank given the association with different amounts of information across resolutions.  Figure \ref{fig:resolution} shows the detection performance (F1 scores) across the above resolution settings and differing ranks from 15-35, using 10\% seed labels in the belief propagation step.

\par Our experiments suggest that F1 score does increase slightly with higher resolutions and decomposition ranks, but the increases are not significant. We hypothesize that the invariance to changes in resolution is due to the fact that coarse-grained features like number of ads, positions of images in the article, and the overall format of the writing is  still captured even at lower resolutions and the detection is not heavily reliant on the fine-grained features of the articles as shown in Figure \ref{fig:transformation}.  
This finding is promising, as it suggests valuable practical advantages in achieving high performance (88\% F1 score) even using very low resolution or even icon size images and significant associated computational benefits.  Thus, unless specified, in further experiments, we use $200\times100$ images.

\subsection{Investigating Sensitivity to Image Transformation}
Next, we investigate different image-level Transformation to evaluate performance under such settings. Firstly, we consider the importance of colors in the creation of latent patterns and the role they play in the classification task via grayscaling. Next, we explore how vectorizing the channels of color screenshots improves the performance.
\par  We first try to convert the color screenshots into grayscale ones using the below commonly used formula in image processing tasks \cite{Color-to-Grayscale}:
$$
 P=R\times(299/1000)+G\times(578/1000)+B\times(114/1000)
$$
where P, R, G, and B are grayscale, red, green, and blue pixel values, respectively. Next, we create a 3-mode tensor from all grayscale screenshots and apply \VizFake.

\begin{figure}[tb!]
\centering
\subfigure[F1-Real]{\includegraphics[width = 0.65\linewidth]{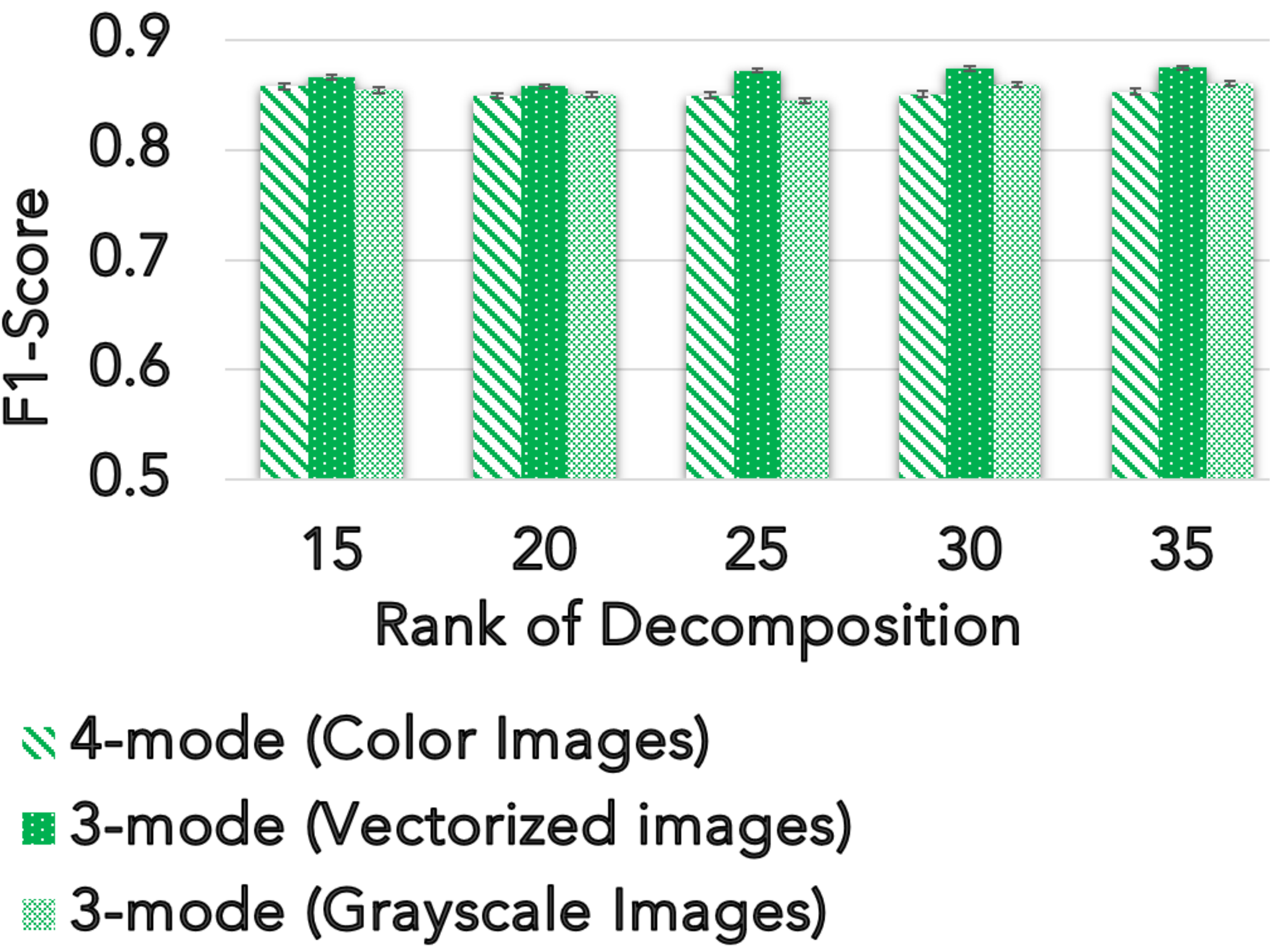}}
\subfigure[F1-Fake]{\includegraphics[width = 0.65\linewidth] {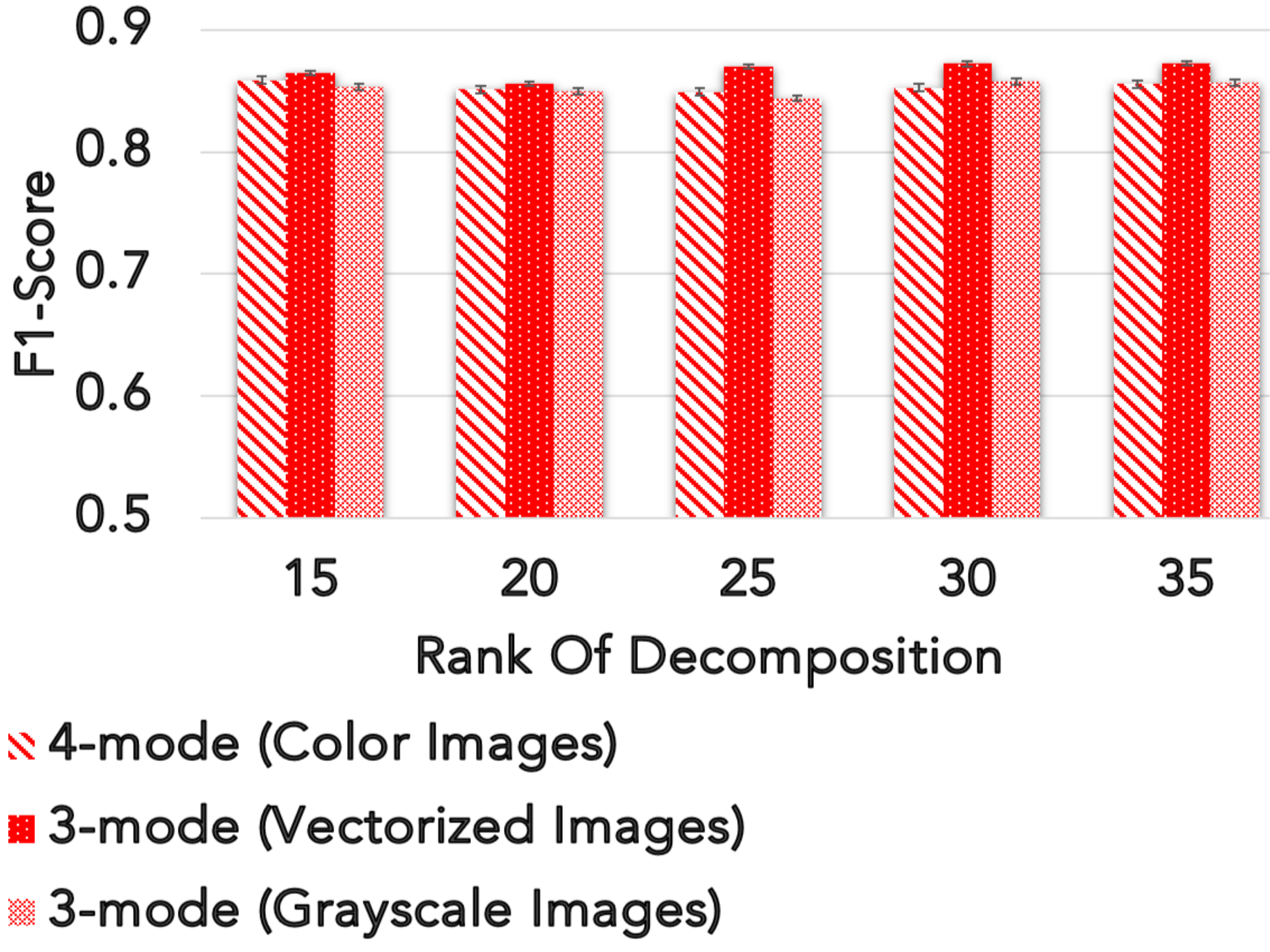}}
\caption{F1 score of 4-mode tensor modeling created out of color screenshots against 3-mode tensors out of vectorized and grayscale screenshots for different ranks.}
\label{fig:grayVscolor2}
\end{figure}

Likewise, to investigate the effect of vectorizing channels of color screenshots, we created another 3-mode tensor by vectorizing each channel matrix. The detection performance using grayscale and vectorized channel tensors in comparison to our standard 4-mode tensor (from color screenshots) are shown in Figure \ref{fig:grayVscolor2}.  Given these different input representations, we again evaluate \VizFake on different rank decompositions.  As shown, in contrast to grayscaling, vectorizing the channels slightly improves the F1 scores. 

\begin{figure}[tb!]
    \centering
    \includegraphics[width = 0.7\linewidth]{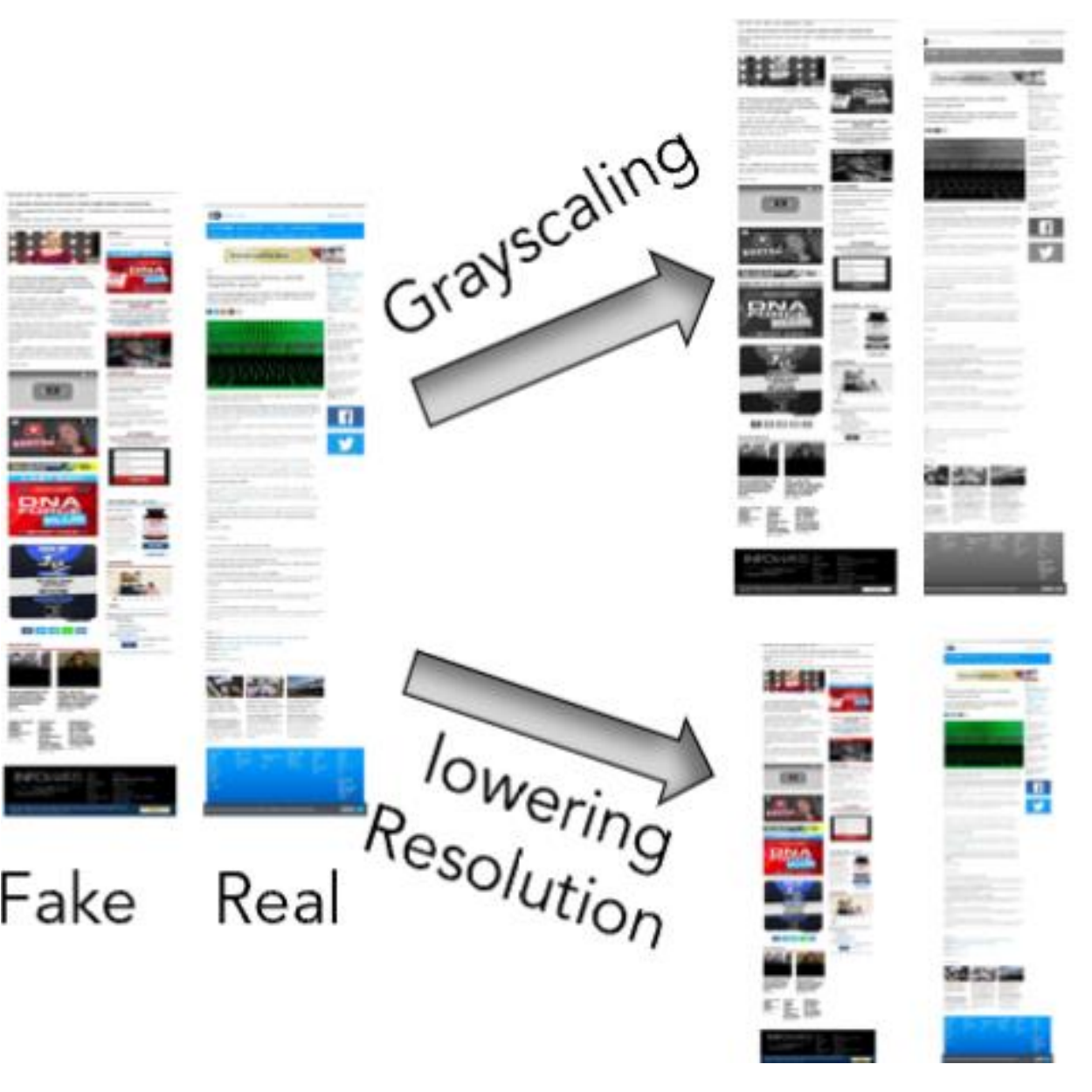}
    \caption{An example of grayscaling and changing the resolution on overall look of screenshots.}
    \label{fig:transformation}
\end{figure}

We hypothesize the rationale for similar grayscale performance to the base 4-mode color model is that several important aspects like number of ads, image positions, writing styles (e.g., number of columns, font) are unaffected and still capture the overall look of the webpage (see Figure \ref{fig:transformation}) and thus producing consistent performance.  The performance improvement for vectorizaing can be explained as follows: By vectorizing an image,
we treat an image as a single observation, or a point in high dimensional pixel space. As a result, we are calculating all possible combinations of pixel statistics, both near and faraway statistics. 
On the other hand, when we consider an image as a matrix, every different image column is treated as an independent observation, and each pixel only covaries with pixels in the rows and the columns and we are not able to capture all possible pairwise statistics. \cite{vasilescu2012multilinear} offers a relevant discussion on vectorization, albeit using subspace arguments rather than latent factor imposed constraints. 
Overall, the minor changes in the F1 score show that \VizFake is robust against common image transformations, suggesting practical performance across various color configurations and image representation schemes.

\subsection{Investigating Sensitivity to Class Imbalance}

\begin{figure}[tb!]
\centering
\subfigure[F1-Real]{\includegraphics[width =0.85\linewidth]{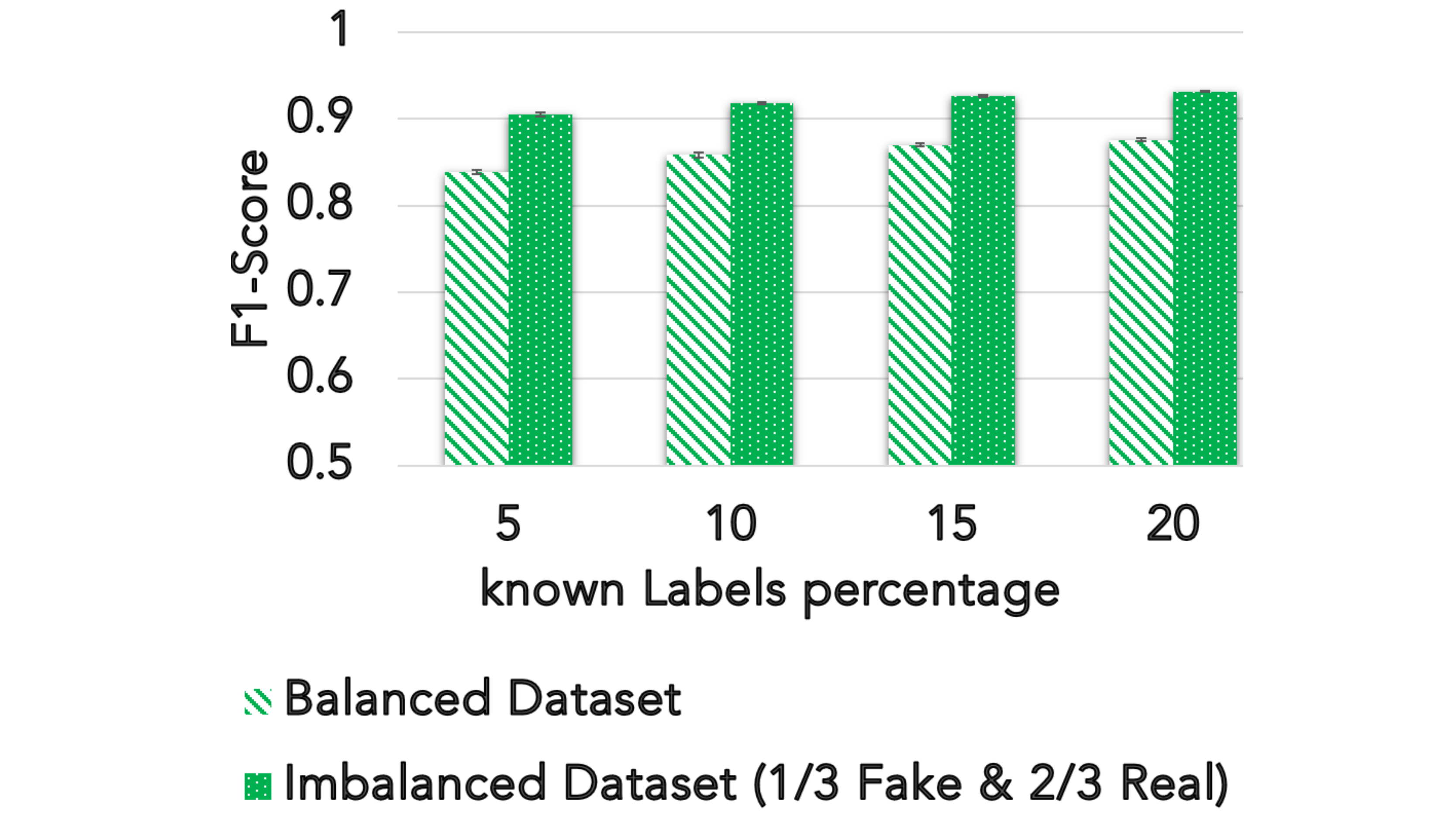}}
\subfigure[F1-Fake]{\includegraphics[width = 0.85\linewidth] {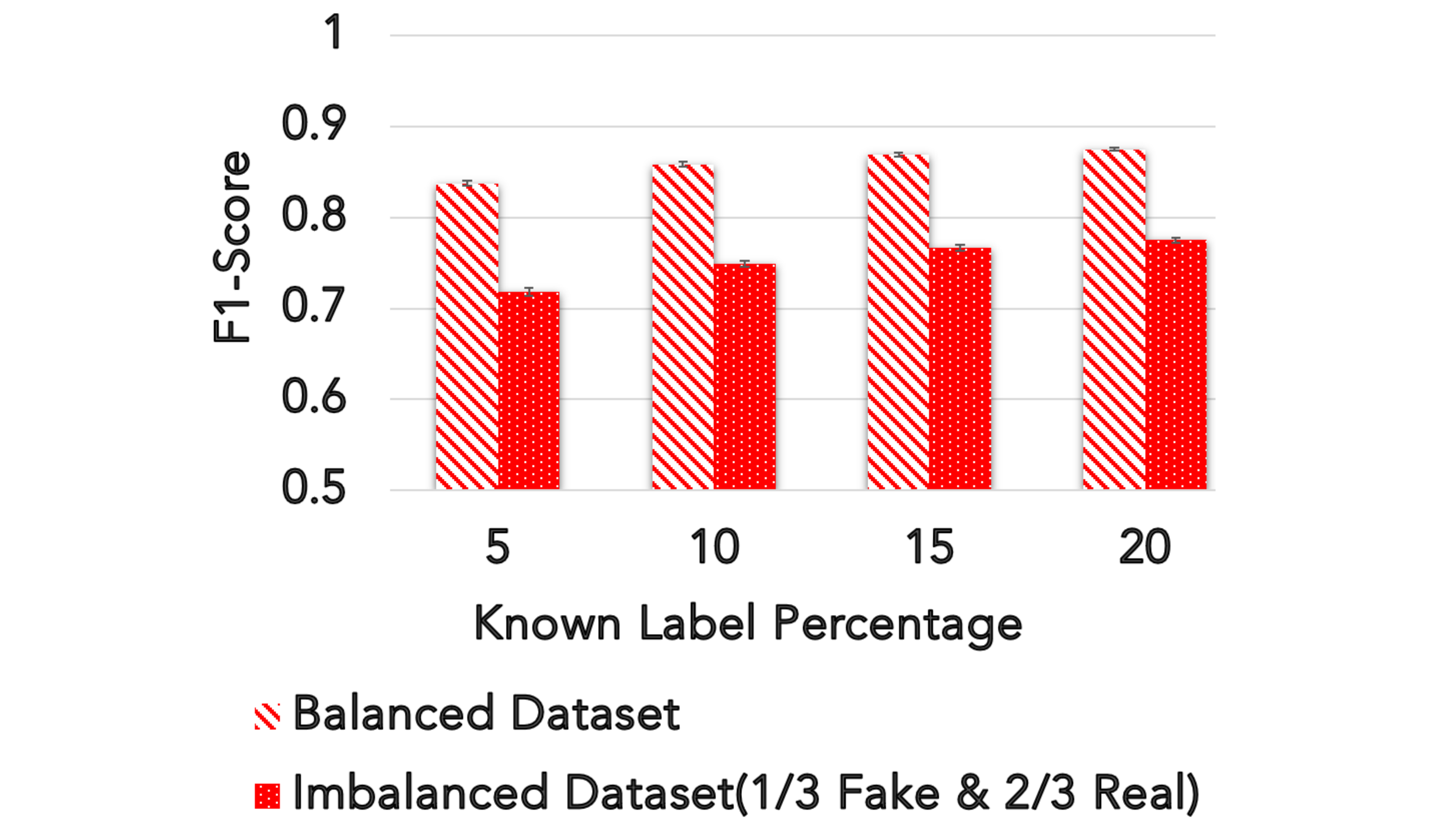}}
\caption{F1 score of using VizFake on an imbalanced dataset (The ratio of screenshots published by fake domains to real ones is $1:2$). On the contrary to fake class, the F1 score of real class increases due to having more samples.}
\label{fig:imbalance}
\end{figure}

Next, we investigate sensitivity of \VizFake to class imbalance, as is often the case in practical settings. We create a dataset of size 50k with a 1:2 fake to real article split. We then assume that the known labels are reflective of the class distribution, and use stratified sampling to designate known labels for the belief propagation step. Figure. \ref{fig:imbalance} shows the F1 scores on both balanced and imbalanced data for different percentages of known labels.

\par As we expect, the F1 score of the fake class drops when we have a scarcity of fake screenshots in the seed label population. Conversely, the F1 score of the real class increases in comparison to a balanced dataset due to more real samples. However, even under the scarcity of fake samples, the F1 score using just $5\%$ of the data is around $70\%$ and using $20\%$ the F1 score is almost $78\%$, suggesting considerably strong results for this challenging task.
Overall, changing the proportion of fake to real articles does expectedly impact classification performance. However, performance on the real class is actually not significantly affected.

\subsection{Investigating Importance of Website Sections}
\begin{figure}[tb!]
    \begin{center}
    \includegraphics[width = 0.8\linewidth]{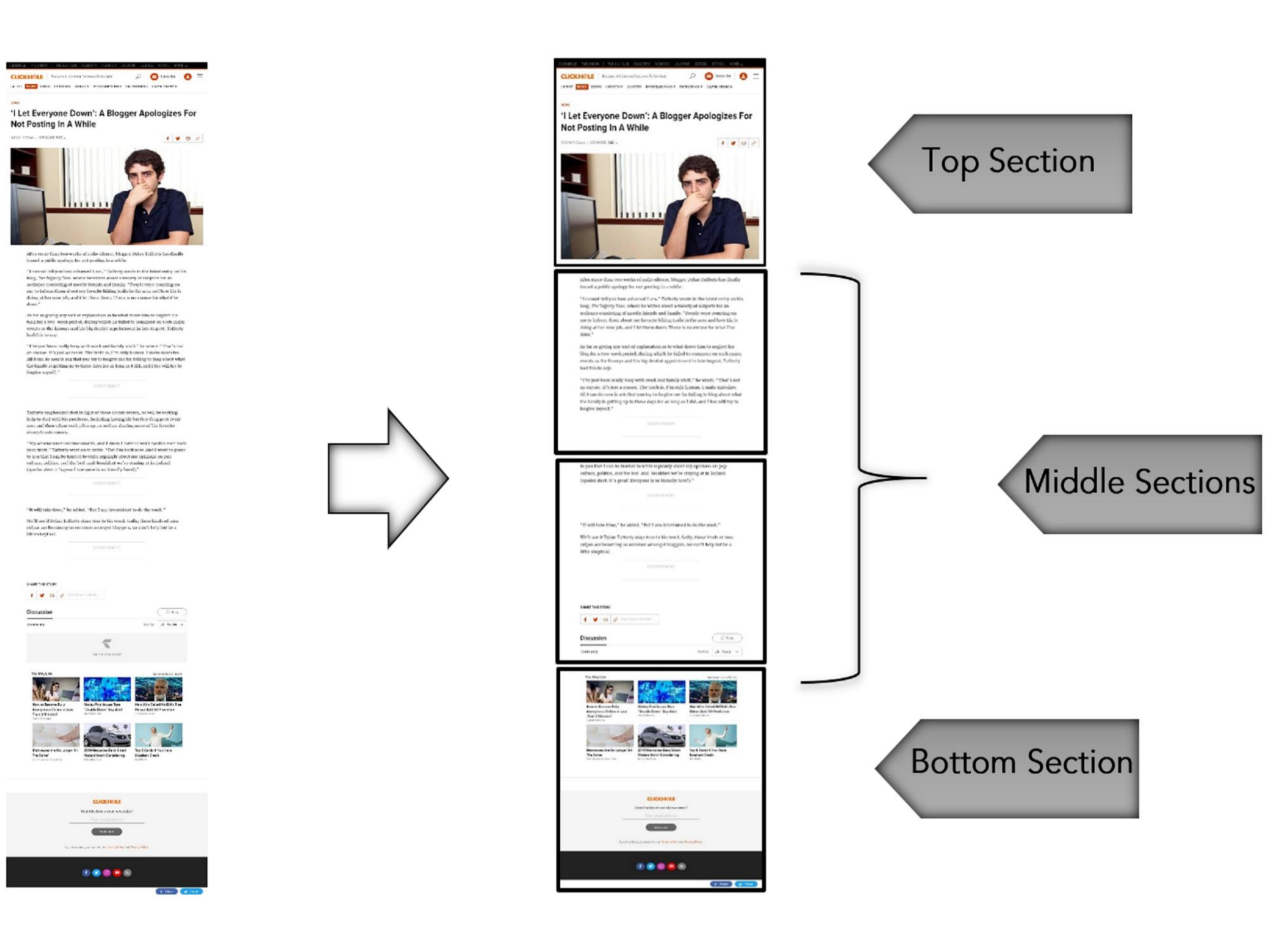}
    \end{center}
    \caption{Cutting a screenshot into four sections.}
    \label{fig:cutting_example}
\end{figure}

\begin{figure}[t!]
\centering
\subfigure[F1-Real]{\includegraphics[width = 0.7\linewidth]{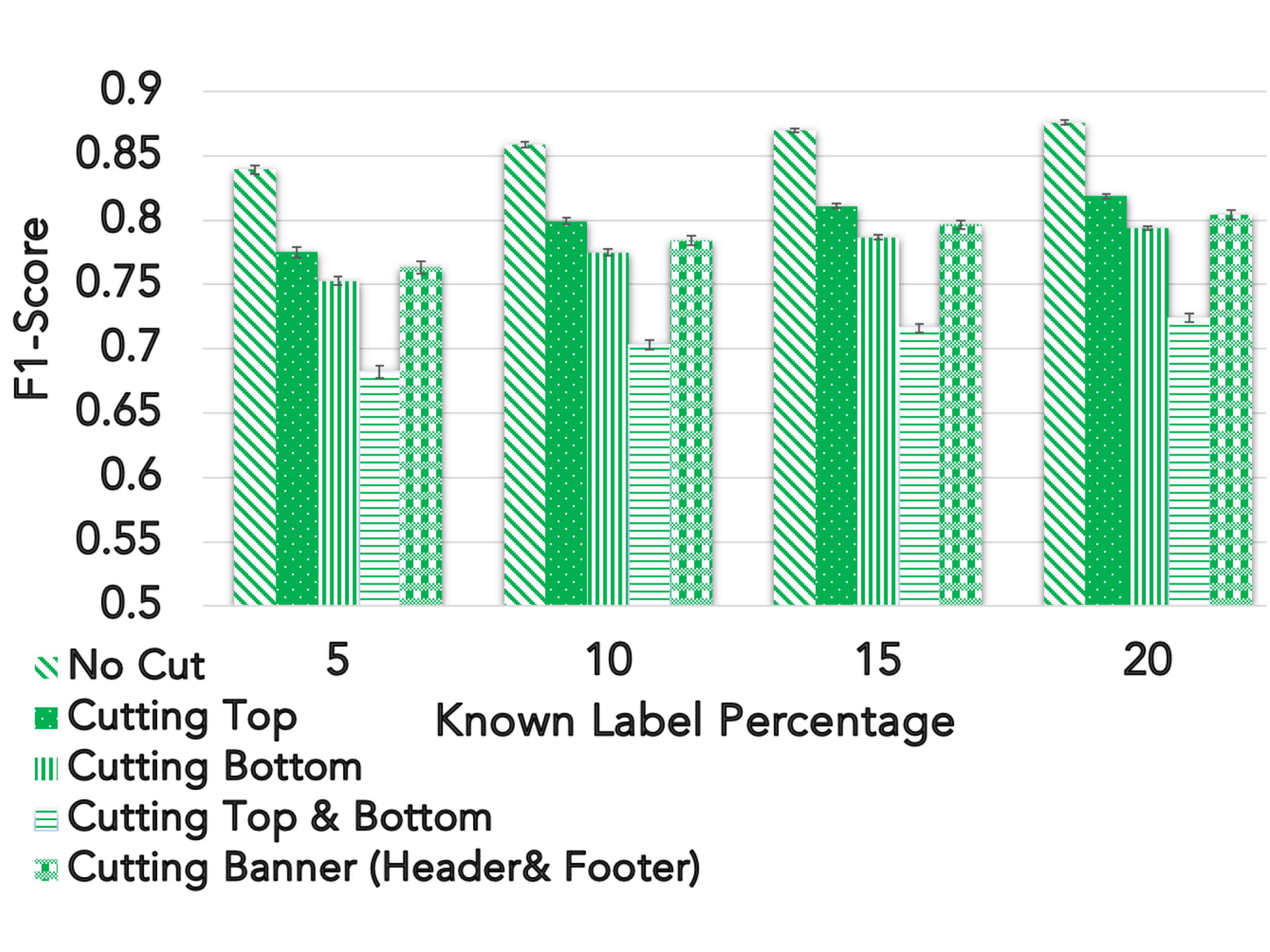}}
\subfigure[F1-Fake]{\includegraphics[width = 0.7\linewidth] {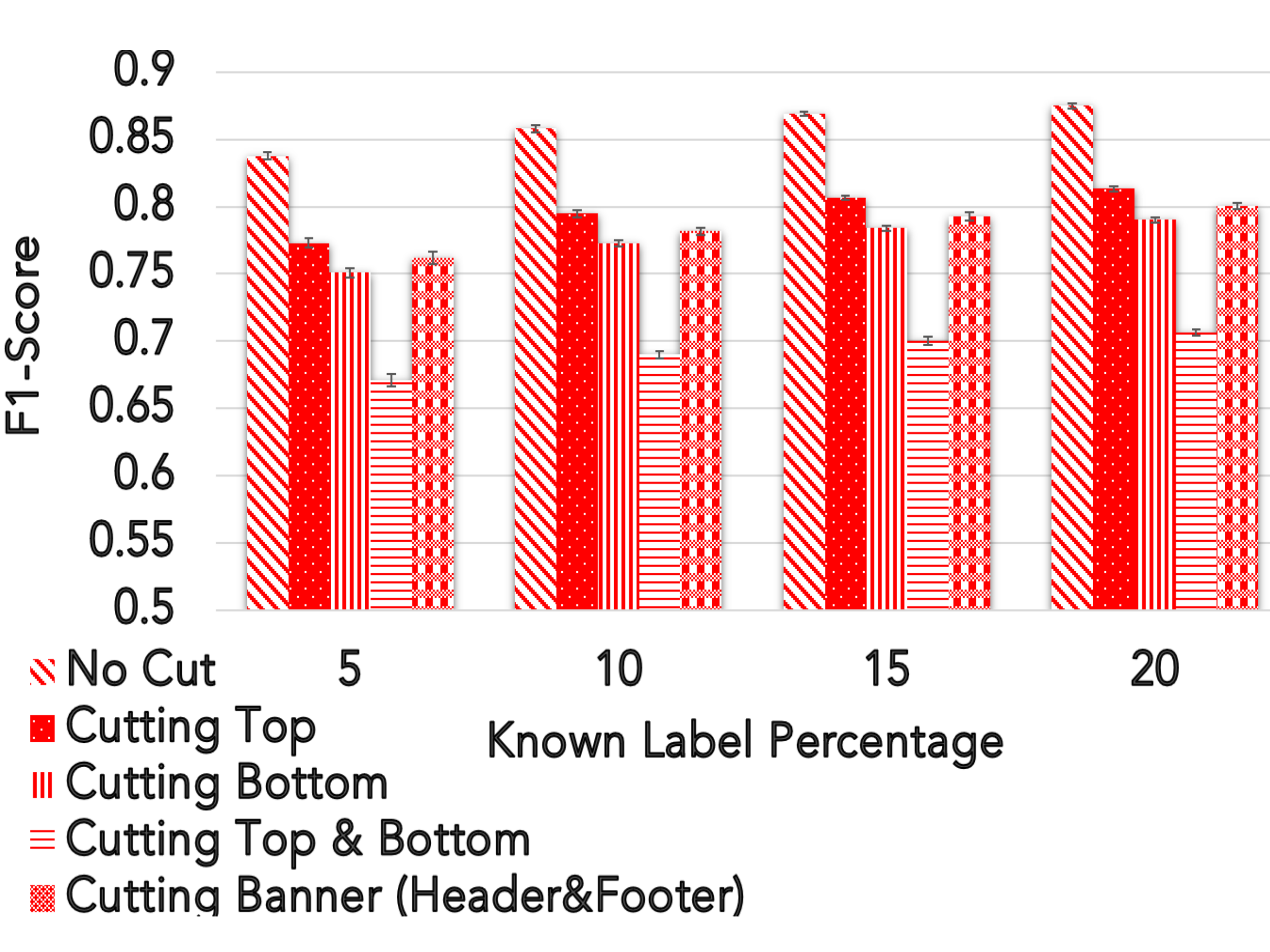}}
\caption{Changes in the F1 score when cutting different sections of the screenshots. In contrast to 2 middle sections, Cutting the top and bottom sections causes a considerable decrease in F1 scores. It seems that style defining style-defining events of the webpages are mostly focused in the top and bottom sections of the webpages.}
\label{fig:cutting}
\end{figure}

One might ask, ``which parts of the screenshots are more informative?'' In other words, in which sections are the latent patterns formed?  To answer these questions, we propose to cut screenshots into four sections as demonstrated in Figure \ref{fig:cutting_example} and use different sections or their combinations while excluding others to create the tensor model (a type of feature ablation study). We propose to create four tensors out of the top, bottom, 2 middle sections, excluding the banner and the concatenation of the top and bottom sections, respectively. For this experiment, we used the 4-mode color tensor and screenshots of size $200\times100$. Thus, each section is of size $50\times100$. Figure \ref{fig:cutting} shows F1 scores of \VizFake on the aforementioned tensors in comparison to using complete screenshots.

\par The results show that by cutting the top or bottom sections of the screenshots the F1 score drops by roughly 6\% and 8\%, respectively. Moreover, if we cut both top and bottom sections the F1 scores decrease significantly by almost 15\%. These two sections convey important information including banners, copyright signatures, sign-in forms, headline images, ads, popups, etc. We noted a considerable portion of the informativeness is included outside the banners, as the banners comprise only 10-20\% of the top/bottom sections and the F1 scores when only excluding the banners are considerably better than when excluding top and bottom both. The middle sections typically consist of the text of the articles, while other article aspects like pictures, ads, and webpage boilerplate tend to be located at the top/bottom sections. Although the top/bottom sections are more informative, the two middle sections still contain important information such as the number of columns, font style, etc. because the middle sections solely, can still classify screenshots with the F1 score of 67\% using just 5\% of labels. By capturing all sections, we achieve significantly stronger results i.e., 83\% F1 using just 5\% labels. This experiment suggests that even if the screenshots are corrupted or censored for privacy considerations e.g., excluding headers and other obvious website tells, we are still capable of identifying fake/real domains using as little as 50\% of the underlying images.

\begin{table*}[tb!]
\centering
\begin{tabular}{c c c c c c c }
\toprule
\multicolumn{7}{c}{\textbf{Fake Class}}\\
\cmidrule(r){2-7}
\multicolumn{1}{c}{}&
\multicolumn{3}{c}{\textbf{VizFake}} 
&\multicolumn{3}{c}{\textbf{VGG16 deep network}}\\
\cmidrule(r){2-4}
\cmidrule(r){5-7}
\textbf{\%labels}&\textbf{F1} &\textbf{Precision}&\textbf{Recall}&\textbf{F1} &\textbf{Precision}&\textbf{Recall}\\
\midrule
5&\textbf{0.852}$\pm$\textbf{0.002}&0.860$\pm$0.005 &0.844$\pm$0.004&0.799$\pm$0.008& 0.823$\pm$0.027&0.779$\pm$0.039\\
10&\textbf{0.871$\pm$0.001}&0.880$\pm$0.003& 0.863$\pm$0.005&0.816$\pm$0.003&0.842$\pm$0.014 &0.793$\pm$0.018\\
15&\textbf{0.881}$\pm$\textbf{0.001}&0.890$\pm$0.002&0.873$\pm$0.003&0.837$\pm$0.001&0.883$\pm$0.009&0.795$\pm$0.009\\
20&\textbf{0.888}$\pm$\textbf{0.001}&0.896$\pm$0.002&0.880$\pm$0.003&0.849$\pm$0.009&0.884$\pm$0.023&0.818$\pm$0.034\\
\bottomrule
\end{tabular}
\caption{VizFake outperforms VGG16 when classifying fake class e.g., F1 score ( $>$ 0.85) with only 5\% of labels.}
\label{table:tensor_model_Fake}
\end{table*}

\begin{table*}[tb!]
\centering
\begin{tabular}{c c c c c c c }
\toprule
\multicolumn{7}{c}{\textbf{Real Class}}\\
\cmidrule(r){2-7}
\multicolumn{1}{c}{}&
\multicolumn{3}{c}{\textbf{VizFake}} 
&\multicolumn{3}{c}{\textbf{VGG16 deep network}}\\
\cmidrule(r){2-4}
\cmidrule(r){5-7}
\textbf{\%labels}&\textbf{F1} &\textbf{Precision}&\textbf{Recall}&\textbf{F1} &\textbf{Precision}&\textbf{Recall}\\
\midrule
5&\textbf{0.854}$\pm$\textbf{0.003}& 0.847$\pm$0.003 &0.862$\pm$0.006&0.809$\pm$0.007&0.790$\pm$0.021&0.830$\pm$0.039\\

10&\textbf{0.874}$\pm$\textbf{0.001}&0.865$\pm$0.004 &0.882$\pm$0.004&0.827$\pm$0.003&0.804$\pm$0.010&0.851$\pm$0.019\\

15&\textbf{0.884}$\pm$\textbf{0.001}&0.876$\pm$0.002& 0.892$\pm$0.003&0.852$\pm$0.002&0.813$\pm$0.005&0.894$\pm$0.010\\

20&\textbf{0.890$\pm$0.001}&0.882$\pm$0.003& 0.898$\pm$0.003&0.860$\pm$0.005&0.831$\pm$0.021&0.892$\pm$0.029\\
\bottomrule
\end{tabular}
\caption{VizFake outperforms VGG16 when classifying real class e.g., F1 score ( $>$ 0.85) with only 5\% of labels.}
\label{table:tensor_model_Real}
\end{table*}
\subsection{Comparing Against Deep-learning Models}
\label{subsec:compare}

A very reasonable first attempt at classification of screenshots, given their wide success in many computer vision tasks, is the use of Deep Convolutional Neural Networks (CNNs). To understand whether or not CNNs are able to capture hidden features that \VizFake scheme cannot extract, we also try CNNs for classification of screenshots. From a pragmatic point of view, we compare i) the classification results each method achieves, and ii) the runtime required to train the model in each case. In what follows,  we discuss the implementation details.

\subsubsection{\VizFake configuration}
We showed that the vectorized tensor outperforms 3-mode grayscale and 4-mode color tensors. So, we choose the 3-mode tensor as tensor model. We use the balanced dataset comprising 50k screenshots with resolution of $200\times100$ and finally we set the rank to 35 based on what is in Figure. \ref{fig:grayVscolor2}. 
\subsubsection{Deep learning configuration}

Although our modest-sized dataset has considerable examples per class (25k), it is not of the required scale for current deep models; thus, we resort to deep transfer learning \cite{pan2009survey}. 

We choose VGG16 \cite{simonyan2014very} pretrained on ImageNet \cite{deng2009imagenet} as our base convolutional network and modify the final fully connected layers to suite our binary classification task.We also tried some other models, they all basically perform similarly. The performance we got was indicative and also was on par with other models. So, we just report the results for VGG16 which is robust enough and the hyper-parameter optimization process is feasible in terms of time and available resources. 
The network is subsequently fine-tuned on screenshot images. 
Due to label scarcity, we want to see if the deep network performs as well as \VizFake when there is a limited amount of labels. Thus, we experiment by fine-tuning on the same label percentage we use for \VizFake. The remaining images are used for validation and testing.

We use the Adam optimizer \cite{kingma2014adam} and search between 0.0001 and 0.01 for the initial learning rate. We apply sigmoid activation in the output layer of the network and the binary cross-entropy as the loss function. The batch sizes we experiment with ranged from 32 to 512 and we finally fixed the batch size for all experiments to 512. Batch size significantly impacts learning as a large enough batch size provides a stable estimate of the gradient for the whole dataset.  \cite{smith2017don,hoffer2017train}. The convergence takes approximately 50 epochs. We note that the effort required to fine-tune a deep network for this task was tedious and included manual trial-and-error, while \VizFake requires the determination of just 2 parameters, both of which produce stable performance across a reasonable range.

\subsubsection{Comparing classification performance}
\label{compare}
Next, we compare the classification performance of \VizFake against the CNN method we explained above in terms of precision, recall, and F1 score. Tables \ref{table:tensor_model_Fake} and \ref{table:tensor_model_Real} show the achieved results of these metrics for \VizFake and CNN model. As demonstrated, \VizFake outperforms CNN especially given less labeled data. For instance, the F1 scores of \VizFake for the fake class when we use only 5\%-10\% of the labels are 85\%-87\%, respectively which is 5-6\% higher than the 80\%-81\% F1 scores from the CNN model. Thus, \VizFake achieves better performance while avoiding considerable time in finding optimal hyperparameters required for tuning VGG16. 

\subsubsection{Comparing the time efficiency}
We evaluate time efficiency by measuring the runtime each method requires to achieve the best results. We experiment on two settings: 

The first one uses a GPU since CNN training is an intensive and time-consuming phase which typically requires performant hardware. Although using a GPU-based framework is not necessary for \VizFake, we re-implemented \VizFake on the same setting we use for the deep learning model to leverage the same scheme, i.e. Python using TensorLy library \cite{kossaifi2019tensorly} with TensorFlow backend. Thus, we avoid influence from factors like programming language, hardware configuration, etc. 

The second configuration uses a CPU and is the one we used in prior experiments and discussed in the Implementation section. Since we are not able to train the CNN model with this configuration due to excessively long runtime, we only report them for \VizFake. 

\par For both experiments, we measure the runtime of bottlenecks, i.e., decomposition of \VizFake and training phase of the deep learning method. Other steps such as: K-NN graph construction, belief propagation, and test phase for CNN method are relatively fast and have negligible runtimes (e.g. construction and propagation for the K-NN graph with 50K screenshots take just 3-4 seconds).  Due to our limited GPU memory, we experiment using a 5\% fraction of the dataset for the GPU configuration. By doing so, we also reduce the I/O overhead that may be counted as execution time when we have to read the dataset in bashes. However, we use 100\% of the dataset for the CPU setting. The technical aspects of each configuration are as follows:

\noindent{\bf Configuration 1:} 
\begin{compactitem}
   \item Keras API for Tensorflow in Python to train the deep network and Python using Tensorly  with TensorFlow backend for \VizFake.
   \item 2 Nvidia Titan Xp GPUs (12 GB)
   \item Training: 5\% (2500 screenshots of size $200\times100$), validation: 4\% (2000 screenshots)
   \item Decomposition: 5\% (2500 screenshots of size $200\times100$)
\end{compactitem}

\noindent{\bf Configuration 2:} 
\begin{compactitem}
   \item Matlab Tensor Toolbox 2.6
   \item CPU: Intel(R) Core(TM) i5-8600K CPU @ 3.60GHz
   \item Decomposition: 100\% (50K screenshots)
\end{compactitem}
\par The average number of iterations, time per iteration, and average total time for 10 runs of both methods on Configuration 1 and the same metrics for \VizFake on Configuration 2 are reported in Tables  \ref{table:Config1} and \ref{table:Config2}, respectively.

\par Based on execution times demonstrated in Table \ref{table:Config1}, the tensor-based method is roughly 216 and 31.5 times faster than the deep learning method in terms of average time and average time per iteration, respectively. Moreover,  the iterations required for \VizFake is almost 7 times less than the epochs required for the CNN method. Note that these results are very conservative estimates since we do not consider time spent tuning CNN hyperparameters in this evaluation. Table \ref{table:Config2} shows the execution time for \VizFake on Configuration 2. Decomposing a tensor of 50k color screenshots using CPU is roughly 3 Mins for screenshots of size $200\times100$, increasing to 6 Mins for larger tensors.
  
\begin{table}[tb!]
\centering
\begin{tabular}{c c c c}
\toprule
\textbf{Resolution}&\textbf{Avg.\# Iter.}&\textbf{Avg. Time/Iter. } &\textbf{Avg. Time}\\
\midrule
$200 \times 100$ & $7.64$ &$23.76$s & $181.55$s\\
$300 \times 100$ & $7.88$ & $35.52$s &$279.95$s\\
$400 \times 100$ & $7.72$ & $47.82$s &$369.22$s\\
\bottomrule
\end{tabular}
\caption{Execution time (Sec.) of VizFake  for different resolutions on configuration 2}
\label{table:Config2}
\end{table}

\begin{table}[tb!]
\centering
\begin{tabular}{c c c c}
\toprule
\textbf{Method}&\textbf{Avg. \# Iter.}&\textbf{Avg. time/Iter. } &\textbf{Avg. Time}\\
\midrule
VizFake & \textbf{7.08} & \textbf{1.05s} & \textbf{7.64s}  \\
\midrule
CNN  &50 & $33.08$s & $1654$s \\
\bottomrule
\end{tabular}
\caption{Execution times (Sec.) of VizFake and CNN deep learning model on configuration 1.}
\label{table:Config1}
\end{table}
\par Overall, the results suggest that \VizFake is \emph{2 orders of magnitude faster} than the state-of-the-art deep transfer learning method for the application at hand, and generally more ``user-friendly'' for real-world deployment.

\begin{table*}[tb!]
\centering
\begin{tabular}{ c c c c c c}
\toprule
\multicolumn{6}{c}{\textbf{Fake Class}}\\
\cmidrule(r){2-6}
\textbf{\%labels} & \textbf{TF-IDF/SVM}& \textbf{Doc2Vec/SVM}&\textbf{GloVe/LSTM}& \textbf{FastText}&\textbf{VizFake}
\\
\toprule
5&
0.812$\pm$0.005&0.511$\pm$0.000& 0.651$\pm$0.019&0.717$\pm$0.010&\textbf{0.844$\pm$0.004}\\

10&0.828$\pm$0.001&0.530$\pm$0.004&0.672$\pm$0.024&0.748$\pm$0.007&\textbf{0.863$\pm$0.005}\\

15&0.836$\pm$0.002&0.540$\pm$0.004& 0.699$\pm$0.020&0.757$\pm$0.006&\textbf{0.873$\pm$0.003}\\

20&0.841$\pm$0.001&0.546$\pm$0.002&0.718$\pm$0.002&0.758$\pm$0.004&\textbf{0.880$\pm$0.003}\\
\bottomrule
\end{tabular}
\caption{The F1 score of VizFake for fake class, outperforms the F1 score of state of the art text-based approaches.}
\label{table:text_based_Fake}
\end{table*}
\begin{table*}[tb!]
\centering
\begin{tabular}{ c c c c c c}
\toprule
\multicolumn{6}{c}{\textbf{Real Class}}\\
\cmidrule(r){2-6}
\textbf{\%labels} & \textbf{TF-IDF/SVM}& \textbf{Doc2Vec/SVM}&\textbf{GloVe/LSTM}& \textbf{FastText}&\textbf{VizFake}
\\
\toprule
5&0.814$\pm$0.004&0.511$\pm$0.000&0.650$\pm$ 0.028& 0.650$\pm$ 0.030&\textbf{0.862$\pm$0.006}\\

10
&0.829$\pm$0.005&0.520$\pm$0.001&0.680$\pm$0.005&0.707$\pm$ 0.016&\textbf{0.882$\pm$0.004}\\

15
&0.836$\pm$0.003&0.526$\pm$0.002&0.698$\pm$0.013&0.712$\pm$0.010&\textbf{0.892$\pm$0.003}\\

20
&0.842$\pm$0.001&0.534$\pm$0.006&  0.712$\pm$0.009&0.728$\pm$0.009&\textbf{0.898$\pm$0.003}\\
\bottomrule
\end{tabular}
\caption{The F1 score of VizFake for real class, outperforms the F1 score of state of the art text-based approaches.}
\label{table:table:text_based_Real}
\end{table*}
\subsection{Comparing Against Text-based Methods}
Even though the main goal of this work is to explore whether or not we can leverage the overall look of the serving webpage to discriminate misinformation, we compare the classification performance of \VizFake with some well-known text-based approaches to investigate how successful is the proposed approach in comparison to these widely used methods. We compare against:
\begin{itemize}
 \item {\textbf{\tfidf}} term frequency–inverse document frequency method is one of the widely used methods for document classification. \tfidf models the importance of words in documents.  We create a \tfidf model out of screenshots text and then we leverage SVM for classification.
    \item \textbf{{\docvec}} a shallow 2-layers neural network proposed by Google \cite{doc2vec}. \docvec is an extension to word2vec and generate vectors for documents. Again, we use SVM classifier.\footnote{\tiny
   {https://github.com/seyedsaeidmasoumzadeh/Binary-Text-Classification-Doc2vec-SVM}}
    \item \textbf{{\fasttext}} a proposed NLP library by Facebook Research. \fasttext learns the word representations which can be used for text classification. It is shown that the accuracy of \fasttext is comparable to deep learning models but is considerably faster than deep competitors\footnote{\tiny{https://github.com/facebookresearch/fastText}}\cite{bojanowski2016enriching}.
   \item \textbf{{\GloVeLSTM}} a linear vector representation of the words using an aggregated global word-word co-occurrence. We create a dictionary of unique words and leverage Glove to map indices of words into a pre-trained word embedding\cite{lin-al-2017-embed-iclr}. Finally, we leverage a LSTM classifier\footnote{\tiny{https://github.com/prakashpandey9/Text-Classification-Pytorch}} pre-trained on IMDB and fine-tune it on our dataset. We examined embedding length in range 50-300 and finally set it to 300. The tuned batch size and hidden size are 256, 64 respectively.
\end{itemize} 
 The experimental results of the aforementioned methods are given in Table. \ref{table:text_based_Fake} and Table. \ref{table:table:text_based_Real}. As demonstrated, the classification performance of \VizFake reported in these tables, outperforms the performance of the shallow network approaches i.e., \docvec and \fasttext as well as the deep network approach i.e., \GloVeLSTM which shows the capability of \VizFake in comparison to neural network methods in settings that there is a scarcity of labels. The \tfidf representation along with SVM classifier leads to classification performance close to the proposed visual approach which illustrates that visual information of the publishers is as discriminative as the best text-based approaches.

\subsection{Comparing Against Website Structure Features}
A question that may come to mind is "why not using website features instead of screenshots?" To address this question, we repeat the proposed pipeline i.e., decomposition, K-NN graph, and belief propagation this time using HTML tags crawled from the serving webpages. To this end, we create an article/tags matrix then we decompose this matrix using Singular Value Decomposition ($
\mathbf{X} \approx\mathbf{U} \boldsymbol{\Sigma} \mathbf{V}^{T}$) and leverage matrix $\mathbf{U}$ which corresponds to articles pattern to create a K-NN graph and propagate the labels using FaBP. The result of this experiment is given in Table.\ref{table:Tags}. 
As illustrated in Table.\ref{table:Tags}, using HTML tags is highly predictive which is another justification for using the overall look of the webpages. The question raises now is that "Why not just using website features for capturing the overall look, especially when the classification performance is better?" Here is some reasons for using screenshots instead of website features:
\begin{itemize}
\item HTML source of the domain is not always available or even if we gain access to the source, the page may be generated dynamically and as a result, the features that can be informative are probably non-accessible scripted content. This is why the HTML source of our dataset provided us with features mainly related to the high-level structure of the domain shared between different screenshots.
\item HTML feature extraction requires tedious web crawling and data cleaning processes and is difficult to separate useful features from useless ones. Taking screenshots is easy and can be done fast and online needless to extra resources or expert knowledge for web crawling.
\item Even if we have access to the HTML source and be able to separate useful features in an efficient way, these features do not give us any information about the content of the web events such as images, videos, ads, etc. If we are to conduct article-level labeling or even section level labeling (usually just some part of an article is misinformative) we will miss a lot of useful information when we use HTML features while screenshots capture such details. 
\end{itemize}
Given the reasons above, the screenshots are not only as informative as textual content, but also are preferred over time-consuming and often less informative HTML features.
\begin{table}[tb!]
\centering
\begin{tabular}{c c c c c}
\toprule
\textbf{\%labels} & \textbf{5}& \textbf{10}&\textbf{15}\\
\toprule
 \textbf{Fake}&0.977$\pm$0.0004&0.983$\pm$0.0002&0.985$\pm$0.0002\\
 \midrule
 \textbf{Real}&0.977$\pm$0.0004&0.983$\pm$0.0003&0.985$\pm$0.0002\\
\bottomrule
\end{tabular}
\caption{Performing proposed pipeline on HTML-Tags of articles. The result justifies that HTMLs only contain domain features which is shared between all articles of that domain.}
\label{table:Tags}
\end{table}

\subsection{Exploratory Analysis}
The tensor representation of \VizFake is not only highly predictive in semi-supervised settings, but also lends itself to exploratory analysis, due to the ease of interpretability of the decomposition factors. In this section, we leverage those factors in order to cluster domains into coherent categories (misinformative or not), in an unsupervised fashion. Each column of the screenshot embedding $\mathbf{C}$  indicates the membership of each screenshot to a cluster, defined by each of the rank-one components (for details on how to generally interpret CP factors as clustering, see \cite{Papalexakis:2016}). Each one of the clusters has a representative latent image, which captures the overall intensity in different parts of the image indicating regions of interest that are participating in generating that cluster.
To obtain this image, we compute the outer product of column vectors of matrices corresponding to pixels and channels i.e., $\mathbf{A}$ and $\mathbf{B}$ for the vectorized tensor and scale it to range 0-255 which provides us with $R$ latent images. We then annotate the images based on the ground truth only to verify that the coherent clusters correspond to fake or real examples. We investigate the interpretability of these latent images by taking the 90th percentile majority vote from the labels of articles with high score in that latent factor. The details of clustering approach is demonstrated in Algorithm. \ref{Algorithm 1}. 
\par
\begin{figure*}[!ht]
\centering
\subfigure[Misinformative latent pattern images]{\includegraphics[width = 0.45\linewidth]{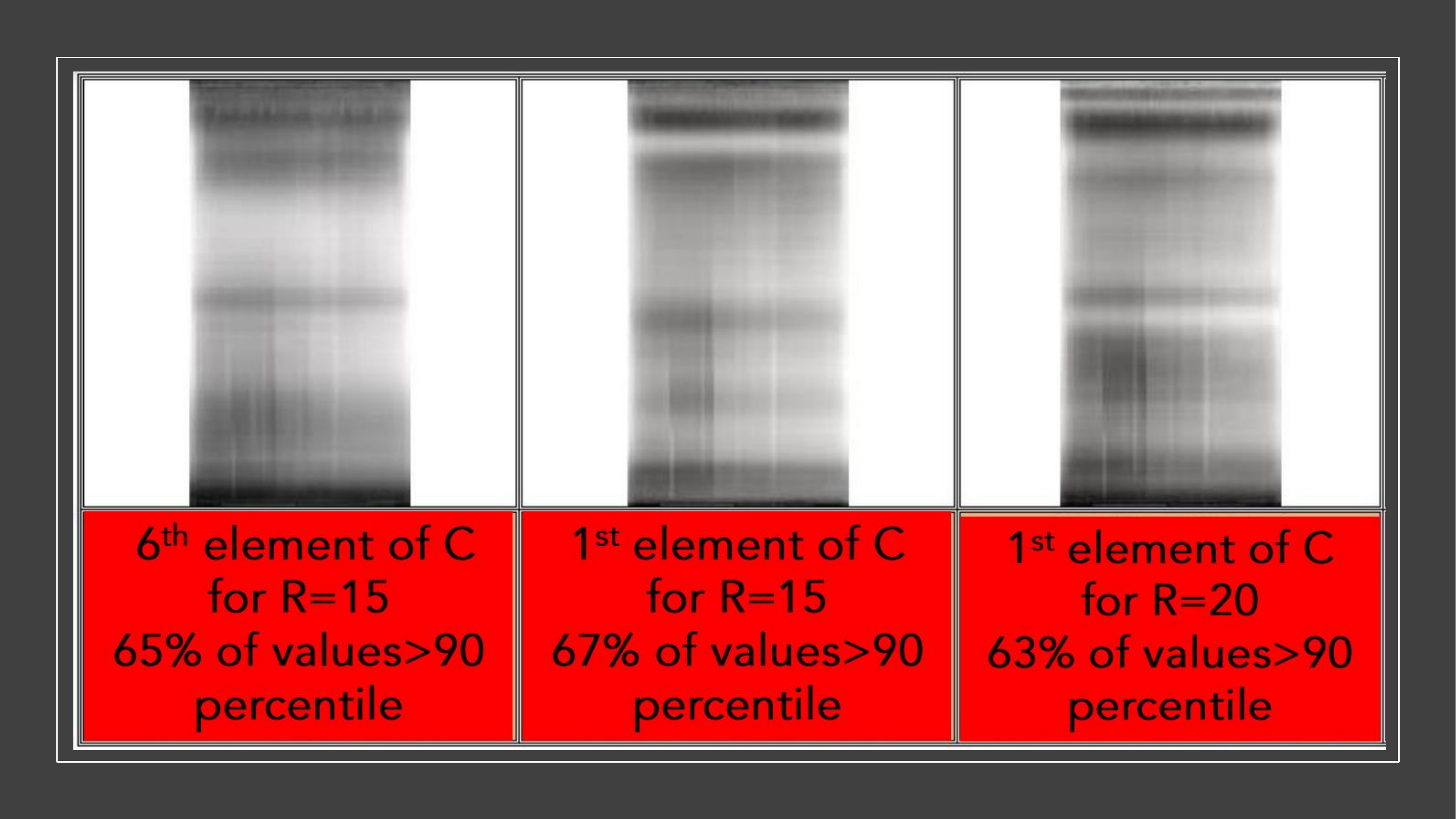}}
\subfigure[Real latent pattern images]{\includegraphics[width = 0.45\linewidth] {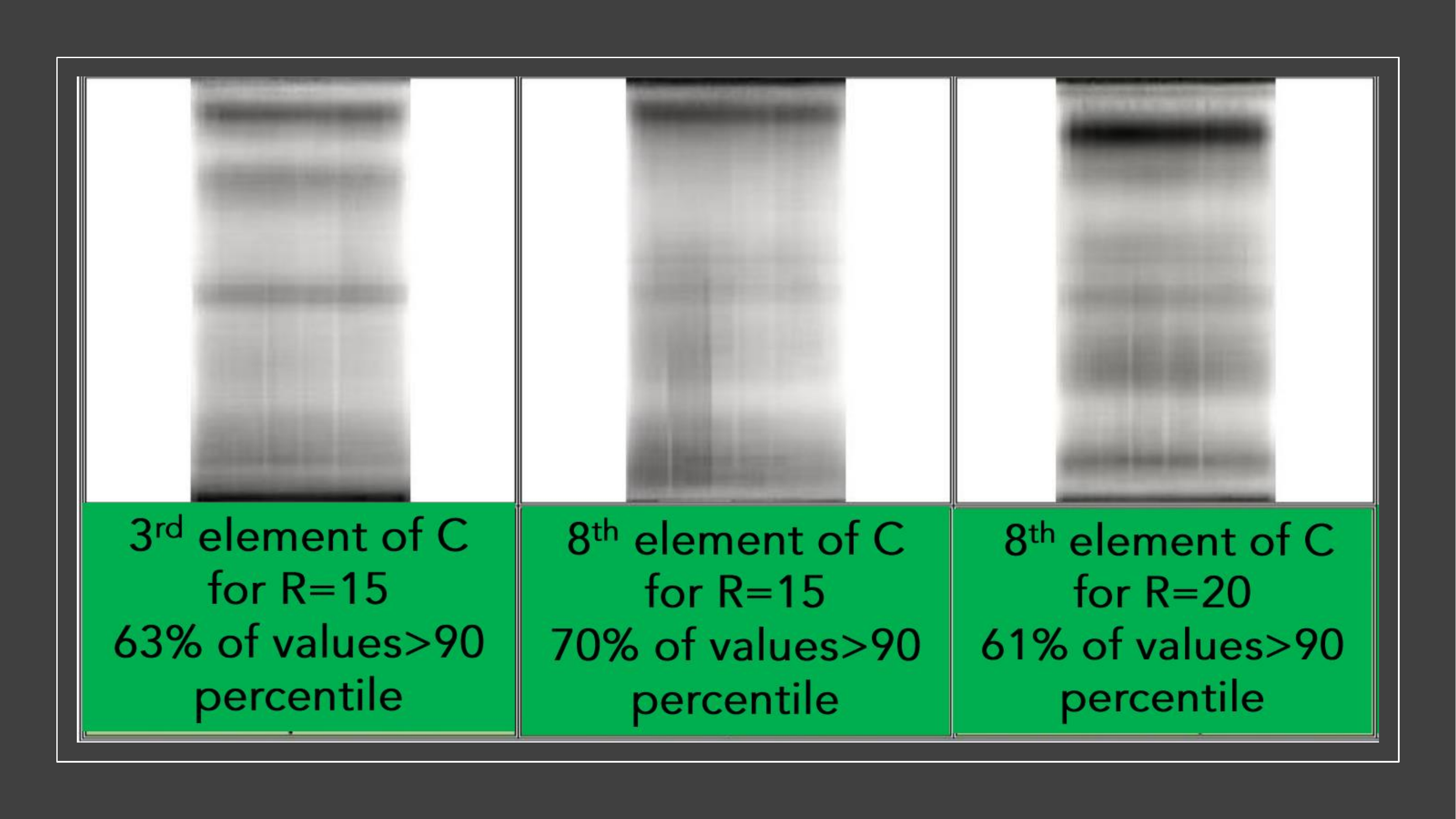}}
\caption{Examples of the cumulative structures of all articles corresponding to factors with the majority of misinformative/real labels. Contrary to the real class, images of misinformative class have darker pixels i.e., the dark portion of the image is wider.}
\label{fig:Samples}
\end{figure*}

 Examples of latent images corresponding to misinformative and real classes are illustrated in Figure \ref{fig:Samples}. The darker a location of an image, the higher degree of ``activity'' it exhibits with respect to that latent pattern. We may view those latent images as ``masks'' that identify locations of interest within the screenshots in the original pixel space.
In Figure \ref{fig:Samples}, we observe that latent images corresponding to real clusters appear to have lighter pixels, indicating little ``activity'' in those locations. For example, the two latent images resulted from rank 15 decomposition are lighter than latent images for the fake class, also the same holds for rank 20. Moreover, as illustrated in Figure \ref{fig:Samples}, darker pixels are more concentrated at the top and the bottom parts of the images which are wider for misinformative patterns and corroborate our assumption about having more objects, such as ads and pop-ups, in fake news websites. As mentioned, such objects are more prevalent at the top and the bottom of the websites which matches our observation here and the cutting observation we discussed earlier. As shown in Figure \ref{fig:cutting}, cutting the bottom and top sections lead to more significant changes in performance than cutting just the banner which also confirms our assumption about informativeness of these sections. This experiment not only provides us with a clustering approach which is obtained without labels and correlates with existing ground truth but also enables us to define filters for misinformation pattern recognition tasks in form of binary masks, that identify locations of interest within a screenshot, which can further focus our analysis.

\subsection{Limitations of the Work}
As discussed earlier, collecting annotation for misinformation detection is a complicated and time-consuming task and as we increase the granularity of the labels from domain level to articles level and even article sections it becomes harder and harder. Moreover, the majority of available ground truth resources like ``BS Detector'' or "NewsGuard" provide labels pertain to domains rather than articles. Despite this disparity, it is shown in several works \cite{helmstetter2018weakly,brief_weakly} that the weakly-supervised task of using labels pertaining to domains, and subsequently testing on labels pertaining to  articles, yields negligible accuracy loss due to the strong correlation between the two targets.
However, as mentioned in the webpage structure section, there are useful article-level information like web events content that can be taken advantage of when we have grainier labels and capturing them causes a drop in performance because they may be considered as noise when working with domain level labels. We defer the study of obtaining and using finer-grained labels for future work.
\begin{algorithm}[tb!]
\SetAlgoLined
\LinesNumberedHidden
\algorithmicrequire{$A$, $B$ and $C$ Factor Matrices}
\\\textbf{Result:} Latent pattern images
\\{$\backslash\backslash$ scale the result to values between 0-255}\\
$min=0; max=255$\\
$a_{ij}=\frac{(a_{ij}-min(a_{ij})\times(max - min)}{(max(a_{ij})-min(a{ij}))} + min$ \\
$b_{ij}=\frac{(b_{ij}-min(b_{ij})\times(max - min)}{(max(b_{ij})-min(b{ij}))} + min$\\
\For{$i=1 \cdots R$}{
$\mathbf{X}_{cumulative}^i \approx \mathbf{a}_i \circ \mathbf{b}_i$
\\${top_n}^i=$ top $(100-\alpha)$ percentile values $c_i$
 \\$\mathbf{X}_{cumulative}^i=$Label-majority-Vote(${top_n}^i$)
}
 \caption{Exploratory analysis} 
 \label{Algorithm 1}
\end{algorithm}
\section{Related Work}
\label{sec:related}
\subsection{Visual-based Misinformation Detection}

The majority of work proposed so far focus on content-based or social-based information. However, there are few studies on visual information of articles. For instance, in \cite{UserImages,Userimage2012} the authors consider user image as a feature to investigate the credibility of the tweets. In another work, Jin et al. \cite{visual2017} define clarity, coherence, similarity distribution, diversity, and visual clustering scores to verify microblogs news, based on the distribution, coherency, similarity, and diversity of images within microblog posts. In \cite{rumors-imagetext} authors find outdated images for the detection of unmatched text and pictures of rumors. Gupta et al. in \cite{fakeimages} classify fake images on Twitter using a characterization analysis to understand the temporal, social reputation of images. On the contrary, we do not focus on the user aspect, i.e., profile image or metadata within a post e.g., image, video, etc. Thus, no matter if there is any images or not, \VizFake captures the overall look of the article.
\subsection{Tensor-based Misinformation Detection}
There are some studies on fake news detection which leverage tensor-based models. For example, in \cite{Hosseinimotlagh2017UnsupervisedCI,ASONAM2018} the authors model content-based information using tensor embedding and try to discriminate misinformation in an unsupervised or semi-supervised regime. In this paper, rather than using article text, we leverage tensors to model article images. Although \VizFake is able to capture the textual look of the article, we are not using time-consuming text analysis and we leverage all features of the article such as text, metadata, domain, etc., when we capture the screenshot of the page.

\section{Conclusions}
\label{sec:conclusions}
In this paper, we leverage a very important yet neglected feature for detecting misinformation, i.e., the overall look of serving domain. We propose a tensor-based model and semi-supervised classification pipeline i.e., \VizFake which outperforms text-based methods and state-of-the-art deep learning models and is over 200 times faster, while also being easier to fine-tune and more practical. Moreover, \VizFake is resistant to some common image transformations like grayscaling and changing the resolution, as well as partial corruptions of the image. Furthermore, \VizFake has exploratory capabilities i.e., it can be used for unsupervised soft-clustering of the articles. \VizFake achieves F1 score of roughly $85\%$ using only $5\%$ of labels for both real and fake classes on a balanced dataset and an F1 score of roughly $95\%$ for real class and $78\%$ for the fake class using only $20\%$ of ground truth on a highly imbalanced dataset.
\section{Acknowledgements}
The authors would like to thank Gisel Bastidas for her invaluable help with data collection. Research was supported by a UCR Regents Faculty Fellowship, a gift from Snap Inc., the Department of the Navy, Naval Engineering Education Consortium under award no. N00174-17-1-0005, and the National Science Foundation CDS\&E Grant no. OAC-1808591. The GPUs used for this research were donated by the NVIDIA Corp. Any opinions, findings, and conclusions expressed in this paper are those of the author(s) and do not necessarily reflect the views of the funding parties.

\end{document}